\def\year{2022}\relax
\documentclass[letterpaper]{article} 
\usepackage{aaai22}  
\usepackage{times}  
\usepackage{helvet}  
\usepackage{courier}  
\usepackage[hyphens]{url}  
\usepackage{graphicx} 
\usepackage{natbib}  
\usepackage{caption} 
\DeclareCaptionStyle{ruled}{labelfont=normalfont,labelsep=colon,strut=off} 
\usepackage{algorithm}
\usepackage{algorithmic}
\usepackage{booktabs}
\usepackage{amsfonts}    
\usepackage{microtype} 
\usepackage{amsmath}
\usepackage{color}
\usepackage{wrapfig,lipsum}
\usepackage{tcolorbox}

\usepackage{enumitem}
\usepackage{adjustbox}
\usepackage{subcaption}
\usepackage{newfloat}
\usepackage{listings}
\floatstyle{ruled}
\newfloat{listing}{tb}{lst}{}
\floatname{listing}{Listing}
\nocopyright
\title{Vision Transformers are Robust Learners}
\author {
    Sayak Paul,\textsuperscript{\rm 1}\equalcontrib~
    Pin-Yu Chen \textsuperscript{\rm 2}\equalcontrib
}
\affiliations {
    \textsuperscript{\rm 1} Carted~
    \textsuperscript{\rm 2} IBM Research\\
    sayak@carted.com, pin-yu.chen@ibm.com
}

\begin{document}

\maketitle

\begin{abstract}
Transformers, composed of multiple self-attention layers, hold strong promises toward a generic learning primitive applicable to different data modalities, including the recent breakthroughs in computer vision achieving state-of-the-art (SOTA) standard accuracy.
What remains largely unexplored is their robustness evaluation and attribution. 
In this work,  we study the robustness of the Vision Transformer (ViT) \cite{dosovitskiy2021an} against common corruptions and perturbations, distribution shifts, and natural adversarial examples. We use six different diverse ImageNet datasets concerning robust classification to conduct a comprehensive performance comparison of ViT \cite{dosovitskiy2021an} models and SOTA convolutional neural networks (CNNs), Big-Transfer \cite{10.1007/978-3-030-58558-7_29}. Through a series of six systematically designed experiments, we then present analyses that provide both quantitative and qualitative indications to explain why ViTs are indeed more robust learners. For example, with fewer parameters and similar dataset and pre-training combinations, ViT gives a top-1 accuracy of 28.10\% on ImageNet-A which is 4.3x higher than a comparable variant of BiT. Our analyses on image masking, Fourier spectrum sensitivity, and spread on discrete cosine energy spectrum reveal intriguing properties of ViT attributing to improved robustness. Code for reproducing our experiments is available at \textcolor{blue}{\url{https://git.io/J3VO0}}. 
\end{abstract}

\section{Introduction}

Transformers \cite{NIPS2017_3f5ee243} are becoming a preferred architecture for various data modalities. This is primarily because they help reduce inductive biases that go into designing network architectures. Moreover, Transformers have been shown to achieve tremendous parameter efficiency without sacrificing predictive performance over architectures that are often dedicated to specific types of data modalities. Attention, in particular, self-attention is one of the foundational blocks of Transformers. It is a computational primitive that allows us to quantify pairwise entity interactions thereby helping a network learn the hierarchies and alignments present inside the input data \cite{DBLP:journals/corr/BahdanauCB14, NIPS2017_3f5ee243}. These are desirable properties to eliminate the need for carefully designed inductive biases to a great extent. 

Although Transformers have been used in prior works \cite{trinh2019selfie, pmlr-v119-chen20s} it was only until 2020, the performance of Transformers were on par with the SOTA CNNs on standard image recognition tasks \cite{carion2020endtoend, touvron2021training, dosovitskiy2021an}. Attention has been shown to be an important element for vision networks to achieve better empirical robustness \cite{hendrycks2021nae}. Since attention is a core component of ViTs (and Transformers in general), a question that naturally gets raised here is -- \textit{could ViTs be inherently more robust?} If so, \textit{why are ViTs more robust learners?} In this work, we provide an affirmative answer to the first question and provide empirical evidence to reason about the improved robustness of ViTs.

Various recent works have opened up the investigation on evaluating the robustness of ViTs \cite{bhojanapalli2021understanding, shao2021adversarial, mahmood2021robustness} but with a relatively limited scope. We build on top of these and provide further and more comprehensive analyses to understand why ViTs provide better robustness for semantic shifts, common corruptions and perturbations, and natural adversarial examples to input images in comparison to SOTA CNNs like Big Transfer (BiT) \cite{10.1007/978-3-030-58558-7_29}. Through a set of carefully designed experiments, we first verify the enhanced robustness of ViTs to common robustness benchmark datasets \cite{DBLP:conf/iclr/HendrycksD19, hendrycks2020many, hendrycks2021nae, xiao2020noise}. We then provide quantitative and qualitative analyses to help understand the reasons behind this enhancement. In summary, we make the following contributions:

\begin{itemize}[leftmargin=*,noitemsep,topsep=0pt,parsep=0pt,partopsep=0pt]
    \item We use 6 diverse ImageNet datasets concerning different types of robustness evaluation and conclude that ViTs achieve significantly better performance than BiTs. 
    \item We design 6 experiments, including robustness to masking, energy/loss landscape analysis, and sensitivity to high-frequency artifacts to study ViT's improved robustness.
    \item Our analysis provides novel insights for robustness attribution of ViT. Moreover, our robustness evaluation and analysis tools are generic and can be used to benchmark and study future image classification models.
\end{itemize}


\section{Related Work}

To the best of our knowledge, \cite{pmlr-v80-parmar18a} first explored the use of Transformers \cite{NIPS2017_3f5ee243} for the task of image super-resolution 
which essentially belongs to the category of image generation. Image-GPT \cite{pmlr-v119-chen20s} used Transformers for unsupervised pre-training from pixels of images. However, the transfer performance of the pre-training method  is not on par with supervised pre-training methods. ViT \cite{dosovitskiy2021an} takes the original Transformers and makes very minimal changes  to make it work with images. In fact, this was one of the primary objectives of ViT i.e. to keep the original Transformer architecture as original as possible and then examining how that pans out for image classification in terms of large-scale pre-training. As noted in \cite{dosovitskiy2021an, steiner2021train}, because of the lesser number of inductive biases, ViT needs to be pre-trained on a relatively larger dataset (such as ImageNet-21k \cite{deng2009imagenet}) with strong regularization for achieving reasonable downstream performance. Strong regularization is particularly needed in the absence of a larger dataset during pre-training \cite{steiner2021train}.

Multiple variants of Transformers have been proposed to show that it is possible to achieve comparable performance on ImageNet-1k \textit{without} using additional data. DeIT \cite{touvron2021training} introduces a novel distillation strategy \cite{44873} to learn a student Transformers-based network from a well-performing teacher network based on RegNets \cite{9156494}. With this approach, DeIT achieves 85.2\% top-1 accuracy on ImageNet-1k without any external data. T2T-ViT \cite{yuan2021tokenstotoken} proposes a novel tokenization method enabling the network to have more access to local structures of the images. For the Transformer-based backbone, it follows a deep-narrow network topology inspired by \cite{zagoruyko2017wide}. With proposed changes, T2T-ViT achieves 83.3\% top-1 accuracy on ImageNet-1k. LV-ViT \cite{jiang2021token} introduces a new training objective namely token labeling and also tunes the structure of the Transformers. It achieves 85.4\% top-1 accuracy on ImageNet-1k. CLIP \cite{radford2021learning} and Swin Transformers \cite{liu2021swin} are also two recent models that make use of Transformers for image recognition problems. In this work, we only focus on ViT \cite{dosovitskiy2021an}.

Concurrent to our work, there are a few recent works that study the robustness of ViTs from different perspectives. In what follows, we summarize their key insights and highlight the differences from our work. \cite{shao2021adversarial} showed that ViTs has better robustness than CNNs against adversarial input perturbations. The major performance gain can be attributed to the capability of learning high-frequency features that are more generalizable and the finding that convolutional layers hinder adversarial robustness. \cite{bhojanapalli2021understanding} studied improved robustness of ViTs over ResNets \cite{7780459} against adversarial and natural adversarial examples as well as common corruptions. Moreover, it is shown that ViTs are robust to the removal of almost any single layer. \cite{mahmood2021robustness} studied adversarial robustness of ViTs through various white-box, black-box and transfer attacks and found that model ensembling can achieve unprecedented robustness without sacrificing clean accuracy against a black-box adversary. This paper shows novel insights that are fundamentally different from these works: (\textbf{i}) we benchmark the robustness of ViTs on a wide spectrum of ImageNet datasets (see Table \ref{tab:dataset-summary}), which are the most comprehensive robustness performance benchmarks to date; (\textbf{ii}) we design six new experiments to verify the superior robustness of ViTs over BiT and ResNet models.

\section{Robustness Performance Comparison on ImageNet Datasets}
\label{perf-comparison}

\subsection{Multi-head Self Attention (MHSA)}

Here we provide a brief summary of ViTs. 
Central to ViT's model design is self-attention \cite{DBLP:journals/corr/BahdanauCB14}. Here, we first compute three quantities from linear projections ($X \in \mathbb{R}^{N \times D}$): (\textbf{i}) \textbf{Q}uery = $X W_{\mathrm{Q}}$,  (\textbf{ii}) \textbf{K}ey = $X W_{\mathrm{K}}$ , and (\textbf{iii}) \textbf{V}alue = $X W_{\mathrm{V}}$, where $W_{\mathrm{Q}}$, $W_{\mathrm{K}}$, and $W_{\mathrm{V}}$ are linear transformations. The linear projections ($X$) are computed from batches of the original input data. Self-attention takes these three input quantities and returns an output matrix ($N \times d$) weighted by attention scores using
\eqref{attn-eq-1}:
\begin{equation}
\label{attn-eq-1}
\operatorname{Attention}(Q, K, V)=\operatorname{Softmax}\left(Q K^{\top} / \sqrt{d}\right) V
\end{equation}
To enable feature-rich hierarchical learning, $h$ self-attention layers (or so-called "heads") are stacked together producing an output of $N \times d h$. This output is then fed through a linear transformation layer that produces the final output of $N \times d$ from MHSA. MHSA then forms the core Transformer block. Additional details about ViT's foundational elements are provided in Appendix \ref{sec_prelim} and \ref{apdnx:patches}.

\subsection{Performance Comparison on Diverse ImageNet Datasets for Robustness Evaluation}

\paragraph{Baselines} In this work, our baseline is a ResNet50V2 model \cite{10.1007/978-3-319-46493-0_38} pre-trained on the ImageNet-1k dataset \cite{russakovsky2015imagenet} except for a few results where we consider ResNet-50 \cite{7780459}\footnote{In these cases, we directly referred to the previously reported results with ResNet-50.}. To study how ViTs hold up with the SOTA CNNs we consider BiT \cite{10.1007/978-3-030-58558-7_29}. At its core, BiT networks are scaled-up versions of ResNets with Group Normalization \cite{Wu_2018_ECCV} and Weight Standardization \cite{qiao2020microbatch} layers added in place of Batch Normalization \cite{pmlr-v37-ioffe15}. Since ViT and BiT share similar pre-training strategies (such as using larger datasets like ImageNet-21k \cite{deng2009imagenet} and JFT-300 \cite{8237359}, longer pre-training schedules, and so on) they are excellent candidates for our comparison purposes. So, a question, central to our work is:
\begin{tcolorbox}
\begin{center}
\textit{Where does ViT stand with respect to BiT in terms of robustness under similar parameter and FLOP regime, pre-training setup, and data regimes, and how to attribute their performance difference?}    
\end{center}
\end{tcolorbox}

\begin{table}[t]
\centering
\adjustbox{max width=1\columnwidth}{
\begin{tabular}{@{}cccc@{}}
\toprule
\textbf{Variant} &
  \textbf{\begin{tabular}[c]{@{}c@{}}\# Parameters \\ (Million)\end{tabular}} &
  \textbf{\begin{tabular}[c]{@{}c@{}}\# FLOPS \\  (Million)\end{tabular}} &
  \textbf{\begin{tabular}[c]{@{}c@{}}ImageNet-1k \\ Top-1 Acc\end{tabular}} \\ \midrule
ResNet50V2           & 25.6138              & \multicolumn{1}{l}{4144.854528} & 76  \\                 
\midrule
BiT m-r50x1          & 25.549352            & 4228.137                        & 80                   \\
BiT m-r50x3          & 217.31908            & 37061.838                       & 84                   \\
BiT m-r101x1         & 44.54148             & 8041.708                        & 82.1                 \\
BiT m-r101x3         & 387.934888           & 71230.434                       & 84.7                 \\
BiT m-r152x4         & 936.53322            & 186897.679                      & 85.39                \\
\midrule
ViT B-16             & 86.859496            & 17582.74                        & 83.97                \\
ViT B-32             & 88.297192            & 4413.986                        & 81.28                \\
ViT L-16             & 304.715752           & 61604.136                       & 85.15                \\
ViT L-32             & 306.63268            & 15390.083                       & 80.99                \\ \bottomrule
\end{tabular}
}
\vspace{-2mm}
\caption{Parameter counts, FLOPS (Floating-Point Operations), and top-1 accuracy (\%) of different variants of ViT and BiT. All the reported variants were pre-trained on ImageNet-21k and then fine-tuned on ImageNet-1k.}
\label{tab:imagenet-1k}
\vspace{-4mm}
\end{table}




Even though BiT and ViT share similar pre-training schedules and dataset regimes there are differences that are worth mentioning. For example, ViT makes use of Dropout \cite{JMLR:v15:srivastava14a} while BiT does not. ViT is trained using Adam \cite{kingma2014adam} while BiT is trained using SGD with momentum. In this work, we focus our efforts on the publicly available BiT and ViT models only. Later variants of ViTs have used Sharpness-Aware Minimization \cite{foret2021sharpnessaware} and stronger regularization techniques to compensate the absence of favored inductive priors \cite{chen2021vision, steiner2021train}. However, we do not investigate how those aspects relate to robustness in this work.

Table \ref{tab:imagenet-1k} reports the parameter counts and FLOPS of different ViT and BiT models 
along with their top-1 accuracy\footnote{Figure 4 of \cite{10.1007/978-3-030-58558-7_29} and Table 5 of \cite{dosovitskiy2021an} were used to collect the top-1 accuracy scores.} on the ImageNet-1k dataset \cite{russakovsky2015imagenet}. It is clear that different variants of ViT are able to achieve comparable performance to BiT but with lesser parameters. 

In what follows, we compare the performance of ViT and BiT on six robustness benchmark datasets \cite{DBLP:conf/iclr/HendrycksD19, hendrycks2020many, hendrycks2021nae}, as summarized in Table \ref{tab:dataset-summary}.
These datasets compare the robustness of ViT, BiT and the baseline ResNet50V2 in different perspectives, including (\textbf{i}) common corruptions, (\textbf{ii}) semantic shifts, (\textbf{iii}) natural adversarial examples, and (\textbf{iv}) out-of-distribution detection. A summary of the datasets and their purpose is presented in Table \ref{tab:dataset-summary} for easier reference.


Notably, in these datasets ViT exhibits significantly better robustness than BiT of comparable parameter counts. Section \ref{sec_analysis} gives the attribution analysis of improved robustness in ViT.

\begin{figure*}
\begin{minipage}[t]{.32\textwidth}
\centering
\includegraphics[width=1\columnwidth]{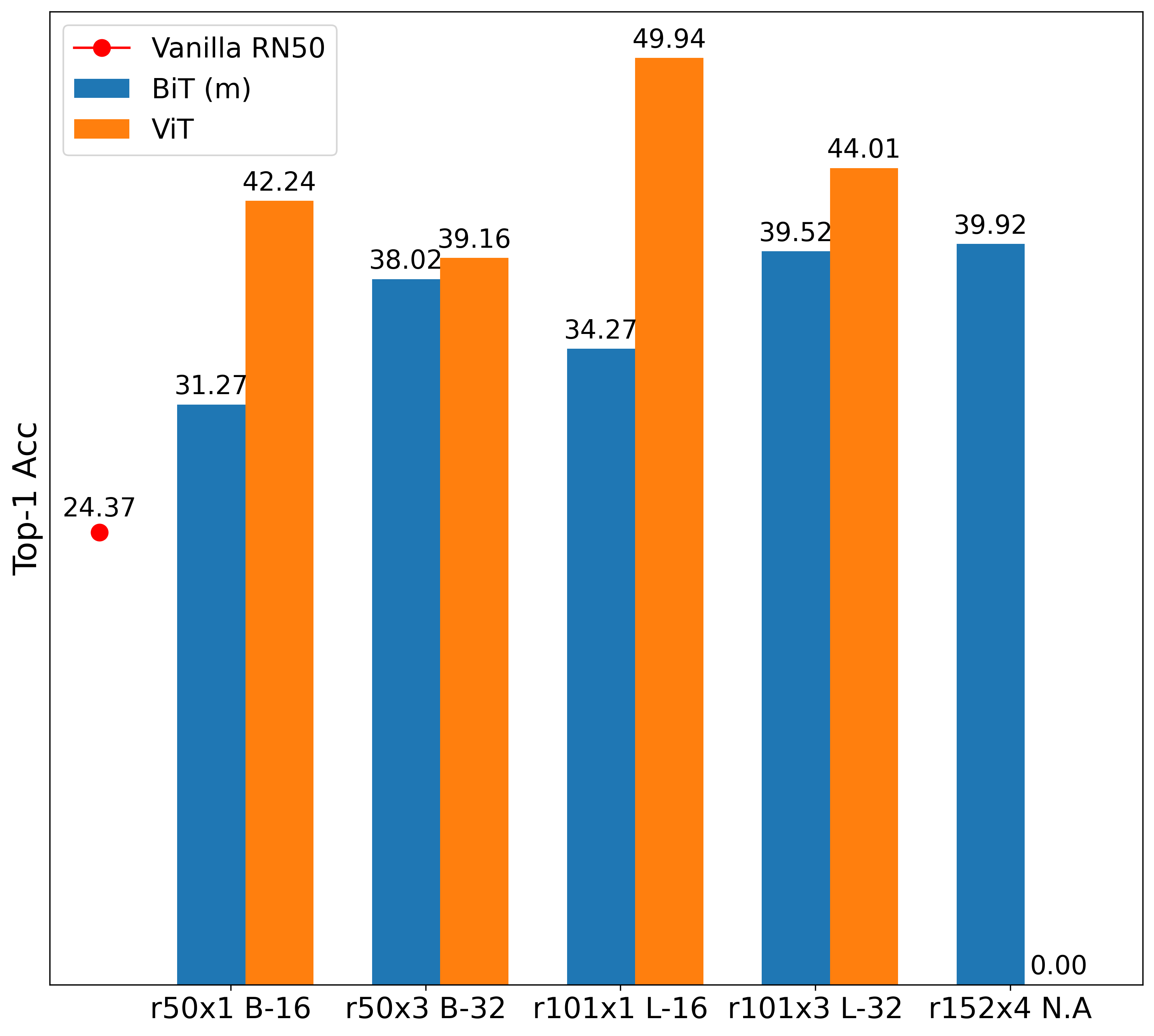} 
\captionof{figure}{Mean top-1 accuracy scores (\%) on the ImageNet-C dataset as yielded by different variants of ViT and BiT.}
\label{fig:imagenet-c}
\end{minipage}
\hfill
    \begin{minipage}[t]{.32\textwidth}
    
    \centering
    \includegraphics[width=1\columnwidth]{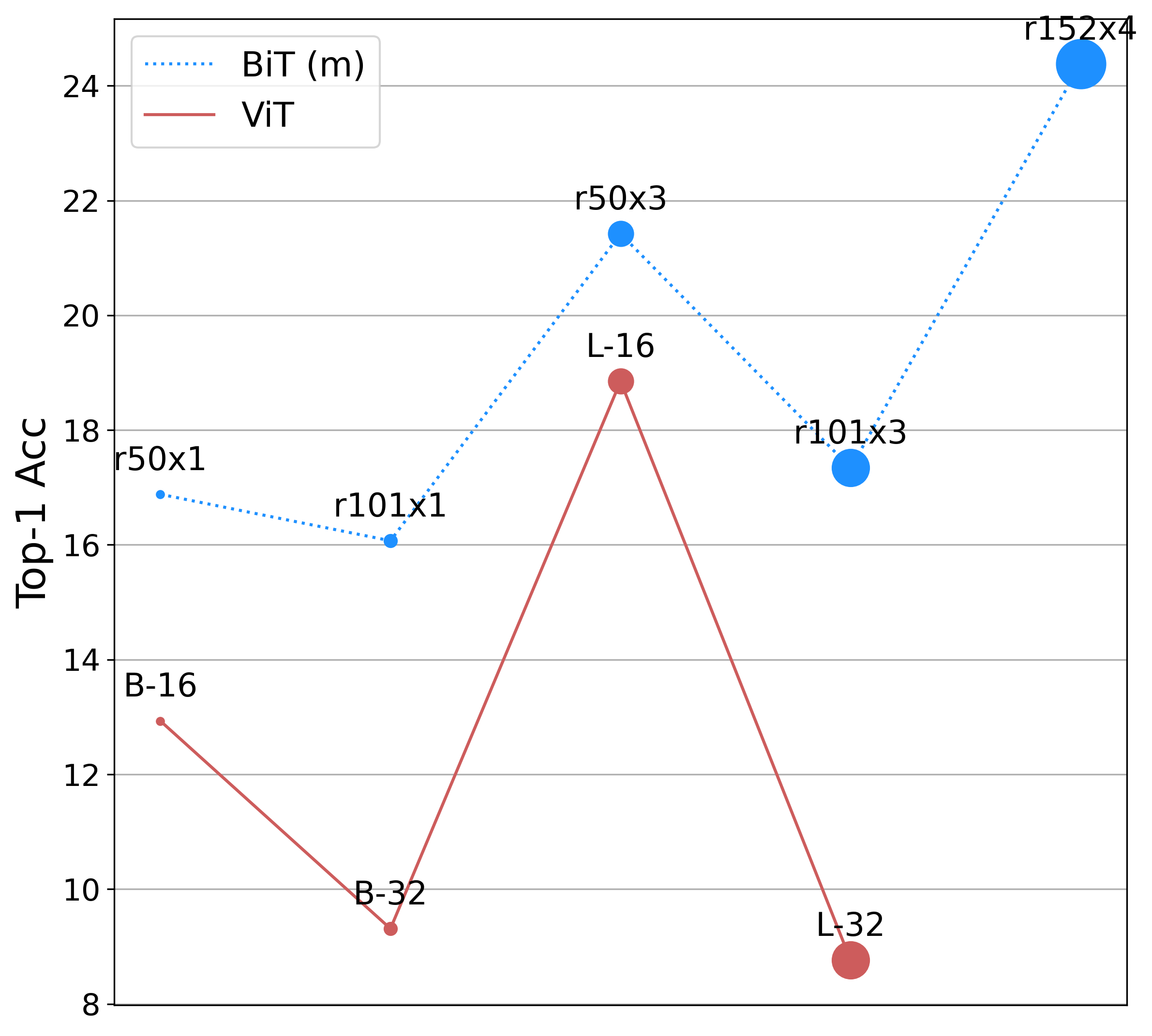} 
    \caption{Top-1 accuracy (\%) of ViT and BiT for contrast corruption (with the highest severity level) on ImageNet-C. 
    }
    \label{fig:imagenet-c-contrast}
    \end{minipage}
\hfill
\begin{minipage}[t]{.32\textwidth}    
\vspace{-50mm}
\centering
    \adjustbox{max width=0.6\textwidth}{
\begin{tabular}{@{}cc@{}}
\toprule
\textbf{Model / Method}         & \textbf{mCE} \\ \midrule
ResNet-50              & 76.7         \\
BiT \texttt{m-r101x3}           & 58.27        \\
DeepAugment+AugMix   & 53.6         \\
ViT L-16           & 45.45        \\
Noisy Student Training  & 28.3         \\ \bottomrule
\end{tabular}
}
\vspace{-2mm}
\captionof{table}{mCEs (\%) of different models and methods on ImageNet-C (lower is better). Note that Noisy Student Training incorporates additional training with data augmentation for noise injection.   }
\label{tab:mce-imagenet-c} 

\adjustbox{max width=1\textwidth}{
\begin{tabular}{@{}ccc@{}}
\toprule
\textbf{Model / Method}        & \textbf{mFR} & \textbf{mT5D} \\ \midrule
ResNet-50             & 58           & 82            \\
BiT-m \texttt{r101x3} & 49.99        & 76.71         \\
AugMix \cite{hendrycks*2020augmix}                & 37.4         & NA            \\
ViT L-16     & 33.064        & 50.15         \\ \bottomrule
\end{tabular}
}
\vspace{-2mm}
\captionof{table}{mFRs (\%) and mT5Ds (\%) on ImageNet-P dataset (lower is better).} 
\label{tab:mfr-imagenet-p}
\end{minipage}
    \vspace{-2mm}    
\end{figure*}

\paragraph{ImageNet-C \cite{DBLP:conf/iclr/HendrycksD19}}
\label{para:imagenet-c}
 consists of 15 types of algorithmically generated corruptions, and each type of corruption has five levels of severity. Along with these, the authors provide additional four types of general corruptions making a total of 19 corruptions. We consider all the 19 corruptions at their highest severity level (5) and report the mean top-1 accuracy in Figure \ref{fig:imagenet-c} as yielded by the variants of ViT and BiT. We consistently observe a better performance across all the variants of ViT under different parameter regimes. Note that BiT \texttt{m-r50x1} and \texttt{m-r101x1} have lesser parameters than the lowest variant of ViT (\texttt{B-16}) but for other possible groupings, variants of ViT have lesser parameters than that of BiT. 
Overall, we notice that ViT performs consistently better across different corruptions except for \textit{contrast}. In Figure \ref{fig:imagenet-c-contrast}, we report the top-1 accuracy of ViT and BiT on the highest severity level of the contrast corruption. This observation leaves grounds for future research to investigate why this is the case since varying contrast factors are quite common in real-world use-cases. Based on our findings, contrast can be an effective but unexplored approach to studying ViT's robustness, similar to the study of human's vision performance \cite{t2013attention,tuli2021convolutional}.

\begin{table}[t]
\centering
\adjustbox{max width=1\columnwidth}{
\begin{tabular}{@{}cc@{}}
\toprule
\textbf{Dataset}                 & \textbf{Purpose}                                                 \\ \midrule
\multicolumn{1}{c}{ImageNet-C \cite{DBLP:conf/iclr/HendrycksD19}} & \multicolumn{1}{c}{\begin{tabular}[c]{@{}c@{}}Common\\ corruptions\end{tabular}}           \\ \midrule
\multicolumn{1}{c}{ImageNet-P \cite{DBLP:conf/iclr/HendrycksD19}} & \multicolumn{1}{c}{\begin{tabular}[c]{@{}c@{}}Common\\ perturbations\end{tabular}}           \\ \midrule
\multicolumn{1}{c}{ImageNet-R \cite{hendrycks2020many}} & \multicolumn{1}{c}{Semantic shifts}                             \\ \midrule
\multicolumn{1}{c}{ImageNet-O \cite{hendrycks2021nae}} & \multicolumn{1}{c}{\begin{tabular}[c]{@{}c@{}}Out-of-domain \\ distribution\end{tabular}}  \\ \midrule
\multicolumn{1}{c}{ImageNet-A \cite{hendrycks2021nae}} & \multicolumn{1}{c}{\begin{tabular}[c]{@{}c@{}}Natural adversarial\\ examples\end{tabular}} \\ \midrule
ImageNet-9  \cite{xiao2020noise}                     & \begin{tabular}[c]{@{}c@{}}Background \\ dependence\end{tabular} \\ \bottomrule
\end{tabular}
}
\vspace{-2mm}
\caption{Summary of the studied datasets and their purpose. }
\label{tab:dataset-summary}
\vspace{-4mm}
\end{table}

In \cite{DBLP:conf/iclr/HendrycksD19}, mean corruption error (mCE) is used to quantify the robustness factors of a model on ImageNet-C. Specifically, the top-1 error rate is computed for each of the different corruption ($c$) types $(1 \leq c \leq 15)$ and for each of the severity ($s$) levels $(1 \leq s \leq 5)$. When error rates for all the severity levels are calculated for a particular corruption type, their average is stored. This process is repeated for all the corruption types and the final value is an average over all the average error rates from the different corruption types. The final score is normalized by the mCE of AlexNet \cite{NIPS2012_c399862d}. 

We report the mCEs for BiT-m \texttt{r101x3}, ViT L-16, and a few other models in Table \ref{tab:mce-imagenet-c}. The mCEs are reported for 15 corruptions as done in \cite{DBLP:conf/iclr/HendrycksD19}. 
We include two additional models/methods in Table \ref{tab:mce-imagenet-c} because of the following: (\textbf{a}) Noisy Student Training \cite{9156610} uses external data and training choices (such as using RandAugment \cite{9150790}, Stochastic Depth \cite{huang2016deep}, etc.) that are helpful in enhancing the robustness of a vision model; (\textbf{b}) DeepAugment and AugMix \cite{hendrycks2020many, hendrycks*2020augmix} are designed explicitly to improve the robustness of models against corruptions seen in ImageNet-C. This is why, to provide a fair ground to understand where BiT and ViT stand in comparison to state-of-the-art, we add these two models. It is indeed interesting to notice that ViT is able to outperform the combination of DeepAugment and AugMix which are specifically designed to provide robustness against the corruptions found in ImageNet-C. As we will discuss in Section \ref{sec_analysis}, this phenomenon can be attributed to two primary factors: 
(\textbf{a}) better pre-training and (\textbf{b}) self-attention. It should also be noted that Noisy Student Training \cite{9156610} incorporates various factors during training such as an iterative training procedure, strong data augmentation transformations from RandAugment for noise injection, test-time augmentation, and so on. These factors largely contribute to the improved robustness gains achieved by Noisy Student Training. 

\paragraph{ImageNet-P \cite{DBLP:conf/iclr/HendrycksD19}}


has 10 types of common perturbations. Unlike the common corruptions, the perturbations are subtly nuanced spanning across fewer number of pixels inside images. As per \cite{DBLP:conf/iclr/HendrycksD19}  mean flip rate (mFR) and mean top-5 distance (mT5D) are the standard metrics to evaluate a model's robustness under these perturbations. They are reported in Table \ref{tab:mfr-imagenet-p}. Since the formulation of mFR and mT5D are more involved than mCE and for brevity, we refer the reader to \cite{DBLP:conf/iclr/HendrycksD19} for more details on these two metrics.
We find ViT's robustness to common perturbations is significantly better than BiT as well as AugMix. 
\begin{figure*}
    \begin{minipage}[t]{.32\textwidth}
\centering
\includegraphics[width=1\columnwidth]{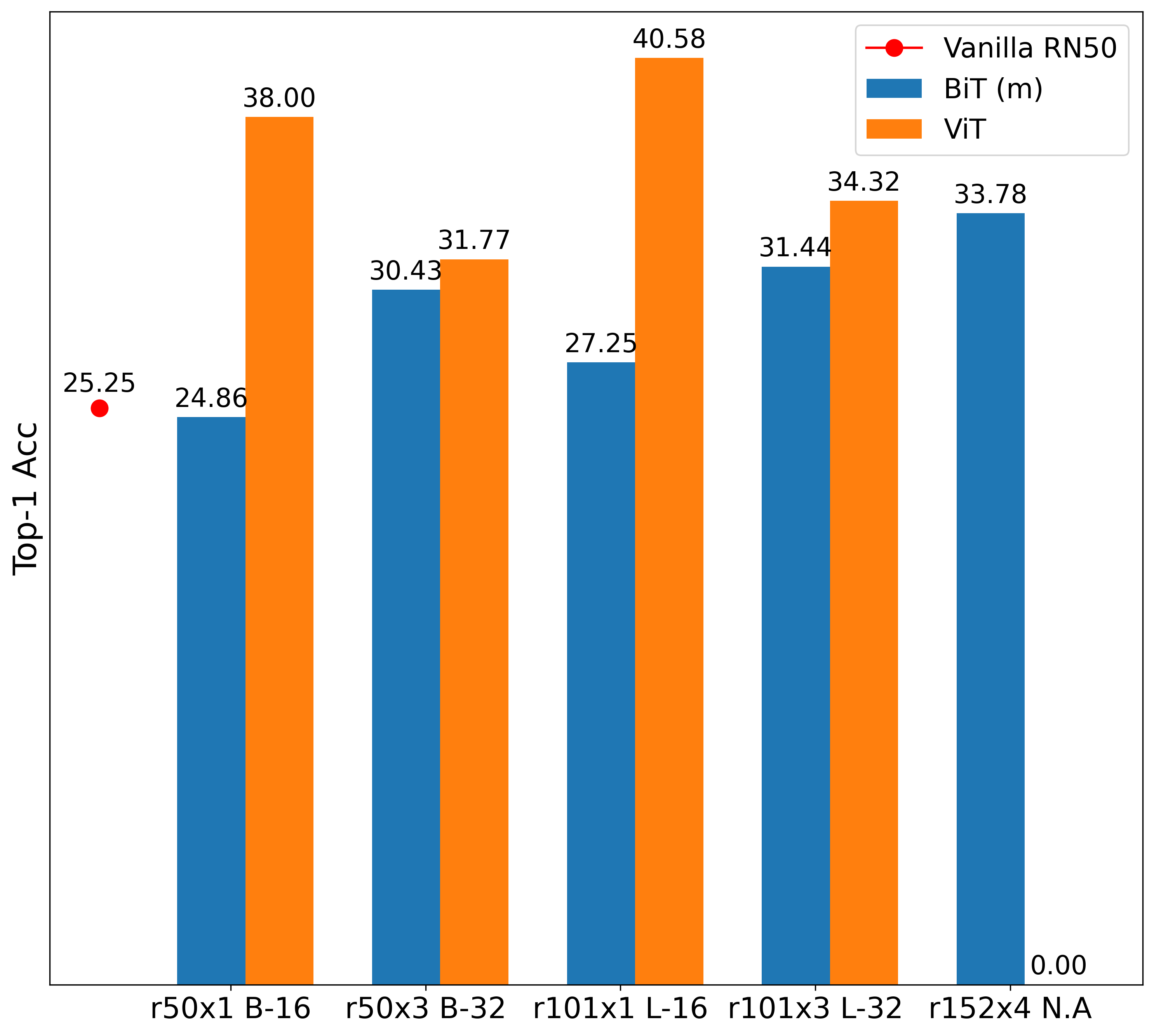} 
\caption{Top-1 accuracy scores (\%) on ImageNet-R dataset. 
}
\label{fig:imagenet-r}
    \end{minipage}
    \hfill
    \begin{minipage}[t]{.32\textwidth}
\centering
\includegraphics[width=1\columnwidth]{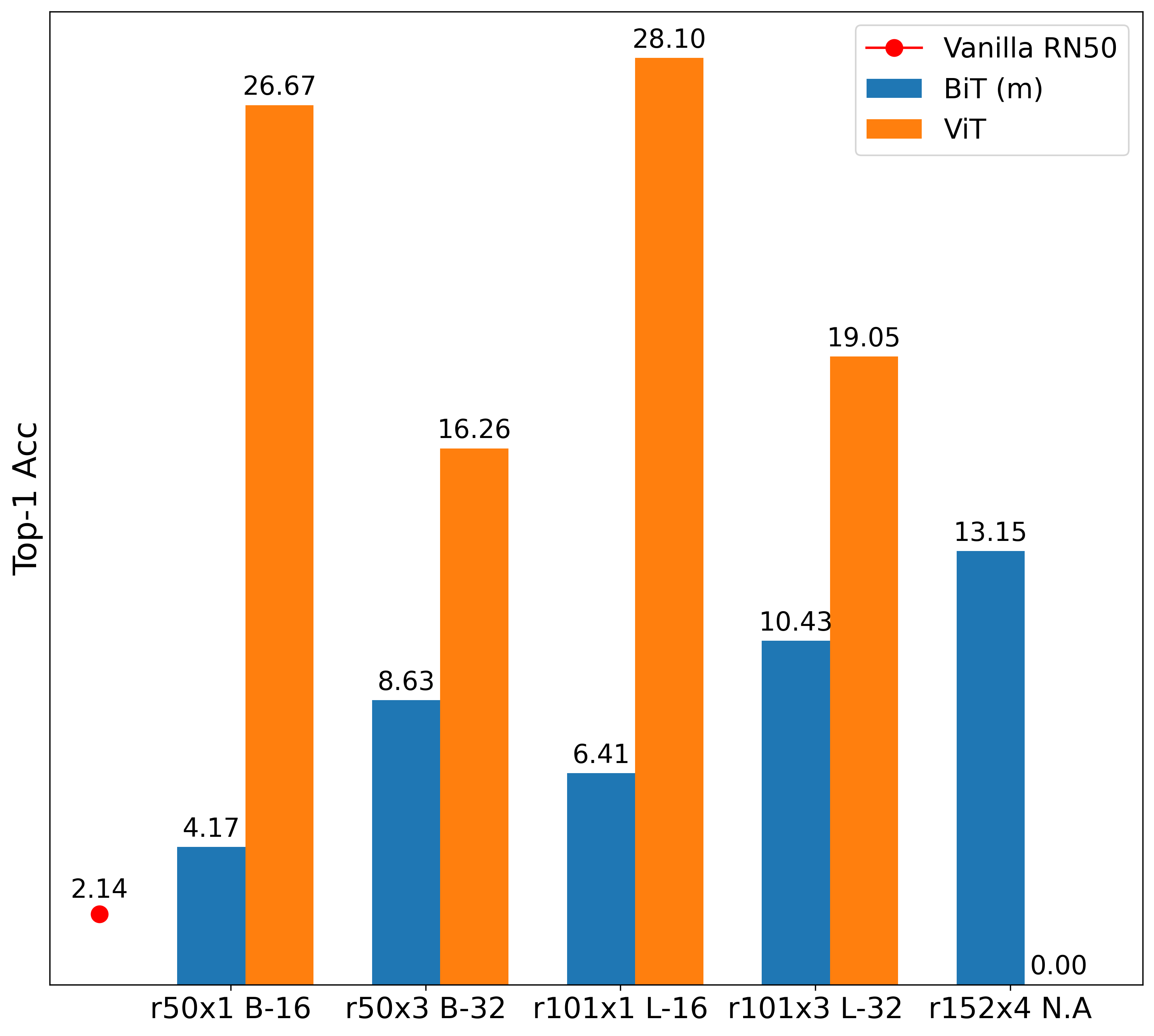} 
\caption{Top-1 accuracy scores (\%) on ImageNet-A dataset. 
}
\label{fig:imagenet-a}
    \end{minipage}
    \hfill
    \begin{minipage}[t]{.32\textwidth}
\centering
\includegraphics[width=1\columnwidth]{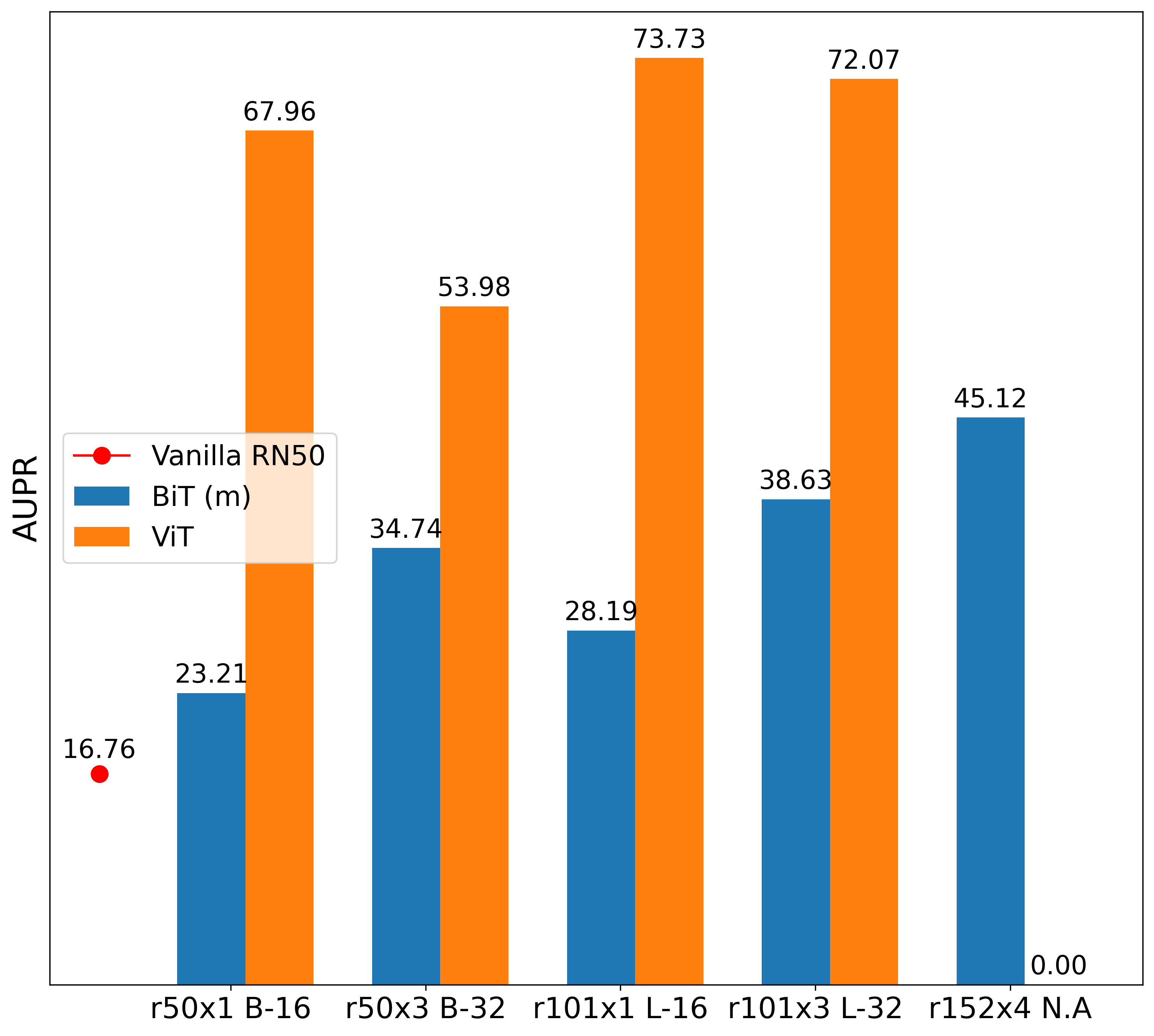} 
\caption{AUPR (higher is better) on ImageNet-O dataset. 
}
\label{fig:imagenet-o}
    \end{minipage}
    \vspace{-2mm}
\end{figure*}

\paragraph{ImageNet-R \cite{hendrycks2020many}}
contains images labelled with ImageNet labels by collecting renditions of ImageNet classes.
It helps verify the robustness of vision networks under semantic shifts under different domains. Figure \ref{fig:imagenet-r} shows that ViT's treatment to domain adaptation is better than that of BiT. 

\paragraph{ImageNet-A \cite{hendrycks2021nae}}

is comprised of natural images that cause misclassifications due to reasons such as multiple objects associated with single discrete categories.  
In Figure \ref{fig:imagenet-a}, we report the top-1 accuracy of ViT and BiT on the ImageNet-A dataset \cite{hendrycks2021nae}. In \cite{hendrycks2021nae}, self-attention is noted as an important element to tackle these problems. This may help explain why ViT performs significantly better than BiT in this case. For example, the top-1 accuracy of ViT L-16
is 4.3x higher than BiT-m \texttt{r101x3}.

\paragraph{ImageNet-O \cite{hendrycks2021nae}}

consists of images that belong to different classes not seen by a model during its training and are considered as \textit{anomalies}. For these images, a robust model is expected to output low confidence scores. We follow the same evaluation approach of using \textit{area under the precision-recall curve} (AUPR) as \cite{hendrycks2021nae} for this dataset. In Figure \ref{fig:imagenet-o}, we report the AUPR of the different ViT and BiT models on the ImageNet-O dataset \cite{hendrycks2021nae}.
ViT demonstrates superior performance in anomaly detection than BiT.

\paragraph{ImageNet-9 \cite{xiao2020noise}}

helps to verify the background-robustness of vision models. In most cases, the foregrounds of images inform our decisions on what might be present inside images. Even if the backgrounds change, as long as the foregrounds stay intact, these decisions should not be influenced. However, do vision models exhibit a similar kind of treatment to image foregrounds and backgrounds? It turns out that the vision models may break down when the background of an image is changed \cite{xiao2020noise}. It may suggest that the vision models may be picking up unnecessary signals from the image backgrounds. In \cite{xiao2020noise} it is also shown that background-robustness can be important for determining models' out of distribution performance. So, naturally, this motivates us to investigate if ViT would have better background-robustness than BiT. We find that is indeed the case (refer to Table \ref{tab:imagenet-9}). Additionally, in Table \ref{tab:imagenet-9-chall}, we report how well BiT and ViT can detect if the foreground of an image is vulnerable\footnote{For details, we refer the reader to the official repository of the background robustness challenge:  \textcolor{blue}{https://git.io/J3TUj}.}. It appears that for this task also, ViT significantly outperforms BiT. Even though we notice ViT's better performance than BiT but it is surprising to see ViT's performance being worse than ResNet-50. We suspect this may be due to the simple tokenization process of ViT to create small image patches that limits the capability to process important local structures \cite{yuan2021tokenstotoken}. 

\section{Why ViT has Improved Robustness?}
\label{sec_analysis}

In this section, we systematically design and conduct six experiments to identify the sources of improved robustness in ViTs from both qualitative and quantitative standpoints.

\begin{table*}[]
  \begin{minipage}[t]{.65\textwidth}
    \centering
    \adjustbox{max width=1\textwidth}{
\begin{tabular}{@{}ccccc@{}}
\toprule
\textbf{Model}           & \textbf{Original} & \textbf{Mixed-Same} & \textbf{Mixed-Rand} & \textbf{BG-Gap} \\ \midrule
BiT-m \texttt{r101x3}    & 94.32           & 81.19            & 76.62             & 4.57          \\ \midrule
ResNet-50    & 95.6           & 86.2           & 78.9             & 7.3          \\ \midrule
ViT L-16 & 96.67  & 88.49    & 81.68    & 6.81 \\ \bottomrule
\end{tabular}
}
\vspace{-2mm}
\captionof{table}{Top-1 accuracy (\%) of ImageNet-9 dataset and its different variants. "BG-Gap" is the gap between "Mixed-Same" and "Mixed-Rand". It measures how impactful background correlations are in presence of correct-labeled foregrounds.}
\label{tab:imagenet-9}  
 \end{minipage} \hfill
   \hspace{2mm}
\begin{minipage}[t]{.3\textwidth}
    \centering
    \adjustbox{max width=0.8\textwidth}{ 
    \begin{tabular}{@{}cc@{}}
\toprule
\textbf{Model}           & \textbf{\begin{tabular}[c]{@{}c@{}}Challenge\\ Accuracy (\%)\end{tabular}} \\ \midrule
BiT-m \texttt{r101x3}    & 3.78                                                                  \\ \midrule
ViT L-16 & 20.02                                                                   \\ \midrule
ResNet-50 & 22.3                                                        \\ \bottomrule
\end{tabular}
}
\vspace{-2mm}
\captionof{table}{Performance on detecting vulnerable image foregrounds from ImageNet-9 dataset.}
\label{tab:imagenet-9-chall}
\end{minipage} 
 \vspace{-4mm}
\end{table*}

\subsection{Attention is Crucial for Improved Robustness}
\label{attention-role}

In \cite{dosovitskiy2021an}, the authors study the idea of ``Attention Distance" to investigate how ViT uses self-attention to integrate information across a given image. Specifically, they analyze the average distance covered by the learned attention weights from different layers. One key finding is that in the lower layers some attention heads attend to almost the entirety of the image and some heads attend to small regions. This introduces high variability in the attention distance attained by different attention heads, particularly in the lower layers. This variability gets roughly uniform as the depth of the network increases. This capability of building rich relationships between different parts of images is crucial for contextual awareness and is different from how CNNs interpret images as investigated in \cite{raghu2021vision}.

Since the attention mechanism helps a model learn better contextual dependencies we hypothesize that this is one of the attributes for the superior performance ViTs show on three robustness benchmark datasets. To this end, we study the performance of different ImageNet-1k models that make use of attention in some form (spatial, channel, or both)\footnote{We used implementations from the \texttt{timm} library for this.}. These models include EfficientNetV2 \cite{pmlr-v139-tan21a} with Global Context (GC) blocks \cite{cao2020global}, several ResNet variants with Gather-Excite (GE) blocks \cite{NEURIPS2018_dc363817} and Selective Kernels (SK) \cite{8954149}. We also include a ViT S/16 model pre-trained on ImageNet-1k for a concrete comparison. We summarize our findings in Table \ref{tab:attn-models}. The results suggest that adding some form of attention is usually a good design choice especially when robustness aspects are concerned as there is almost always a consistent improvement in performance compared to that of a vanilla ResNet-50. This is also suggested by Hendrycks et al. \cite{hendrycks2021nae} but only in the context of ImageNet-A. We acknowledge that the models reported in Table \ref{tab:attn-models} differ from the corresponding ViT model with respect to their training configurations, regularization in particular. But exploring how regularization affects the robustness aspects of a model is not the question we investigate in this work.

Self-attention constitutes a fundamental block for ViTs. So, in a realistic hope, they should be able to perform even better when they are trained in the right manner to compensate for the absence of strong inductive priors as CNNs. We confirm this in Table \ref{tab:attn-models} (last row). Note that the work on AugReg \cite{steiner2021train} showed that it is important to incorporate stronger regularization to train better performing ViTs in the absence of inductive priors and larger data regimes. More experiments and attention visualizations showing the connection between attention and robustness are presented in Appendix \ref{appen_attention}. 

\subsection{Role of Pre-training}

ViTs yield excellent transfer performance when they are pre-trained on larger datasets \cite{dosovitskiy2021an, steiner2021train}. This is why, to better isolate the effects of pre-training with larger data regimes we consider a ViT B/16 model but trained with different configurations and assess their performance on the same benchmark datasets as used in Section \ref{attention-role}. These configurations primarily differ in terms of the pre-training dataset. We report our findings in the Table \ref{tab:pretraining-effects-vit}. We notice that the model pre-trained on ImageNet-1k performs worse than the one pre-trained on ImageNet-21k and then fine-tuned on ImageNet-1k.

Observations from Table \ref{tab:pretraining-effects-vit} lead us to explore another questions i.e., under similar pre-training configurations how do the ViT models stand out with respect to BiT models. This further helps to validate which architectures should be preferred for longer pre-training with larger datasets as far as robustness aspects are concerned. This may become an important factor to consider when allocating budgets and resources for large-scale experiments on robustness. Throughout Section \ref{perf-comparison} and the rest of Section \ref{sec_analysis}, we show that ViT models significantly outperform similar BiT models across six robustness benchmark datasets that we use in this work. We also present additional experiments in Appendix \ref{appen_pre-training} by comparing ViTs to BiTs of similar parameters. 

\begin{table}[t]
\centering
\adjustbox{max width=1.1\columnwidth}{
\begin{tabular}{@{}cccccc@{}}
\toprule
\textbf{Model} &
  \textbf{\begin{tabular}[c]{@{}c@{}}\# Parameters \\ (Million)\end{tabular}} &
  \textbf{\begin{tabular}[c]{@{}c@{}}\# FLOPS \\ (Million)\end{tabular}} &
  \textbf{\begin{tabular}[c]{@{}c@{}}ImageNet-A \\ (Top-1 Acc)\end{tabular}} &
  \textbf{\begin{tabular}[c]{@{}c@{}}ImageNet-R \\ (Top-1 Acc)\end{tabular}} &
  \textbf{\begin{tabular}[c]{@{}c@{}}ImageNet-O \\ (AUPR)\end{tabular}} \\ \midrule
ResNet-50                                                            & 25.6138     & 4144.854528          & 2.14             & 25.25             & 16.76          \\
EfficientV2 (GC)                                                    & 13.678      & 1937.974             & 7.389285         & 32.701343         & 20.34          \\
ResNet-L (GE)                                                       & 31.078      & 3501.953             & 5.1157087        & 29.905242         & 21.61          \\
ResNet-M (GE)                                                       & 21.143      & 3015.121             & 4.99335          & 29.345            & 22.1           \\
ResNet-S (GE)                                                       & 8.174       & 749.538              & 2.4682036        & 24.96156          & 17.74          \\
ResNet18 (SK)                                                       & 11.958      & 1820.836             & 1.802681         & 22.95351          & 16.71          \\
ResNet34 (SK)                                                       & 22.282      & 3674.5               & 3.4683768        & 26.77625          & 18.03          \\
\begin{tabular}[c]{@{}c@{}}Wide (4x)\\  ResNet-50 (SK)\end{tabular} & 27.48       & 4497.133             & 6.0972147        & 28.3357           & 20.58          \\
\textbf{ViT S/16}                                                   & \textbf{22} & \textbf{4608.338304} & \textbf{6.39517} & \textbf{26.11397} & \textbf{22.50} \\ \bottomrule
\end{tabular}
}
\vspace{-2mm}
\caption{Complexity and performance of different attention-fused models on three benchmark robustness datasets. All models reported here operate on images of size 224 $\times$ 224.}
\label{tab:attn-models}
\vspace{-2mm}
\end{table}

\begin{table}[t]
\centering
\adjustbox{max width=0.8\columnwidth}{
\begin{tabular}{@{}cccc@{}}
\toprule
\textbf{Pre-training} &
  \textbf{\begin{tabular}[c]{@{}c@{}}ImageNet-A \\ (Top-1 Acc)\end{tabular}} &
  \textbf{\begin{tabular}[c]{@{}c@{}}ImageNet-R \\ (Top-1 Acc)\end{tabular}} &
  \textbf{\begin{tabular}[c]{@{}c@{}}ImageNet-O \\ (AUPR)\end{tabular}} \\ \midrule
ImageNet-1k &
  8.630994 &
  28.213835 &
  26.25 \\
ImageNet-21k &
  21.746947 &
  41.815233 &
  54.61 \\ \bottomrule
\end{tabular}
}
\vspace{-2mm}
\caption{Performance of the ViT B/16 model on three benchmark datasets.}
\label{tab:pretraining-effects-vit}
\vspace{-2mm}
\end{table}

\begin{table}[t]
\centering
\adjustbox{max width=0.5\columnwidth}{
\begin{tabular}{@{}ccc@{}}
\toprule
\textbf{\begin{tabular}[c]{@{}c@{}}Masking \\ Factor\end{tabular}} &
  \textbf{\begin{tabular}[c]{@{}c@{}}Top-1 \\ Acc (BiT)\end{tabular}} &
  \textbf{\begin{tabular}[c]{@{}c@{}}Top-1 \\ Acc (ViT)\end{tabular}} \\ \midrule
0    & 79   & 83  \\ \midrule
0.05 & 76   & 82.3 \\ \midrule
0.1  & 75   & 81.4  \\ \midrule
0.2  & 72.4 & 77.9  \\ \midrule
0.5  & 52   & 60.4  \\ \bottomrule
\end{tabular}
}
\vspace{-2mm}
\caption{Mean top-1 accuracy (\%) of BiT (\texttt{m-r101x3}) and ViT (\texttt{L-16)} 
with different masking factors.
}
\label{tab:cutout}
\vspace{-4mm}
\end{table}

\begin{table}[t]
\centering
\adjustbox{max width=0.7\columnwidth}{
\begin{tabular}{@{}cccc@{}}
\toprule
\multicolumn{1}{l}{} & \textbf{ResNet-50} & \textbf{BiT-m r101x3} & \textbf{ViT L-16} \\ \midrule
\textbf{P=10}        & 21.8              & 13.9                  & 6.7               \\
\textbf{P=25}        & 30.2              & 14.8                  & 7                 \\
\textbf{P=50}        & 40.4              & 16.4                  & 7.6               \\
\textbf{P=90}        & 58.9              & 23                    & 13.1              \\
\textbf{P=95}        & 63.6              & 24.9                  & 15.1              \\ \bottomrule
\end{tabular}
}
\vspace{-2mm}
\caption{Different percentiles (P) of the error matrix computed from Fourier analysis (Figure \ref{fig:fourier}). }
\label{tab:percentiles-fourier}
\vspace{-4mm}
\end{table}




\subsection{ViT Has Better Robustness to Image Masking}
\label{masking-cutout}



In order to further establish that attention indeed plays an important role for the improved robustness of ViTs, we conduct the following experiment:

\begin{itemize}[leftmargin=*,noitemsep,topsep=0pt,parsep=0pt,partopsep=0pt]
    \item Randomly sample a common set of 1000 images from the ImageNet-1k validation set.
    \item Apply Cutout \cite{devries2017improved} at four different levels: \{5,10,20,50\}\% and calculate the mean top-1 accuracy scores for each of the levels with BiT (\texttt{m-r101x3}) and ViT (\texttt{L-16)}\footnote{We use these two variants because they are comparable with respect to the number model parameters.}. In Cutout, square regions from input images are randomly masked out. It was originally proposed as a regularization technique.
\end{itemize}

Table \ref{tab:cutout} reports that ViT is able to consistently beat BiT when square portions of the input images have been randomly masked out. Randomness is desirable here because ViT can utilize global information. If we fixate the region of masking it may be too restrictive for a ViT to take advantage of its ability to utilize global information. Note that the ViT variant (\texttt{L-16}) we use in this experiment is shallower than the BiT variant (\texttt{m-r101x3}). This may suggest that attention indeed is the strong force behind this significant gain. 

\begin{table*}[]
\begin{minipage}[t]{.49\textwidth}
\centering
\includegraphics[width=1.0\columnwidth]{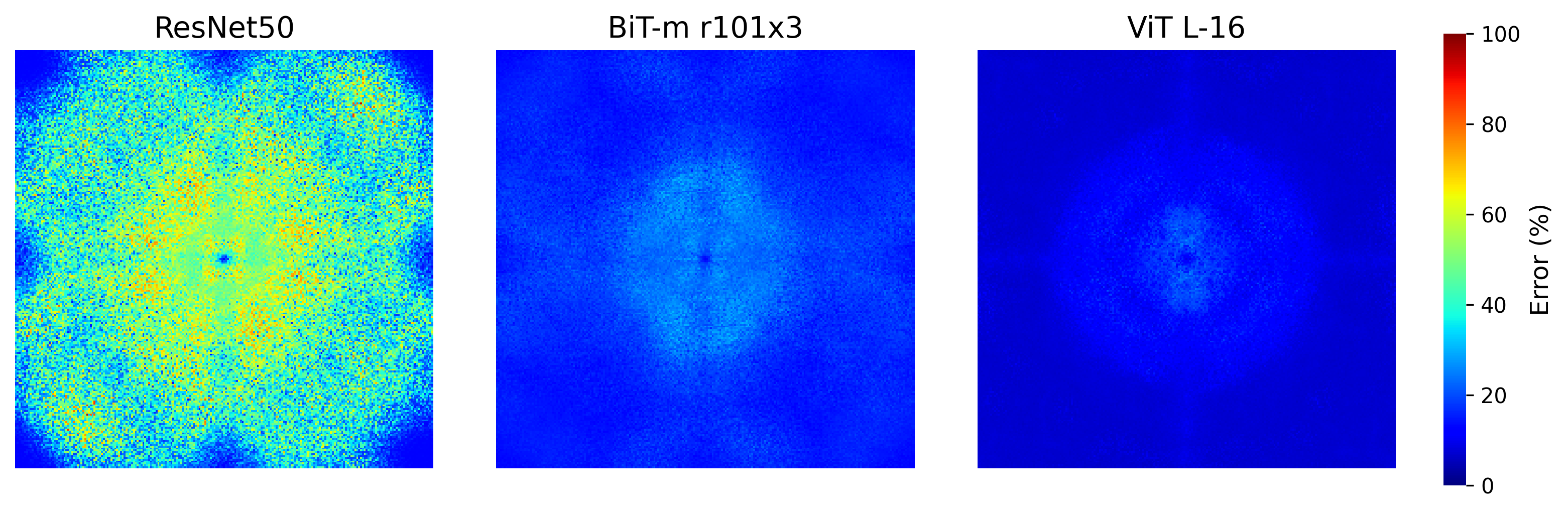} 
\captionof{figure}{Sensitivity heatmap of 2D discrete Fourier
transform spectrum  \cite{NEURIPS2019_b05b57f6}. The low-frequency/high-frequency components are shifted to the center/corner of the spectrum.}
\label{fig:fourier}
\end{minipage}
\hfill
\hspace{1mm}
\begin{minipage}[t]{.49\textwidth}
\centering
\includegraphics[width=1\columnwidth]{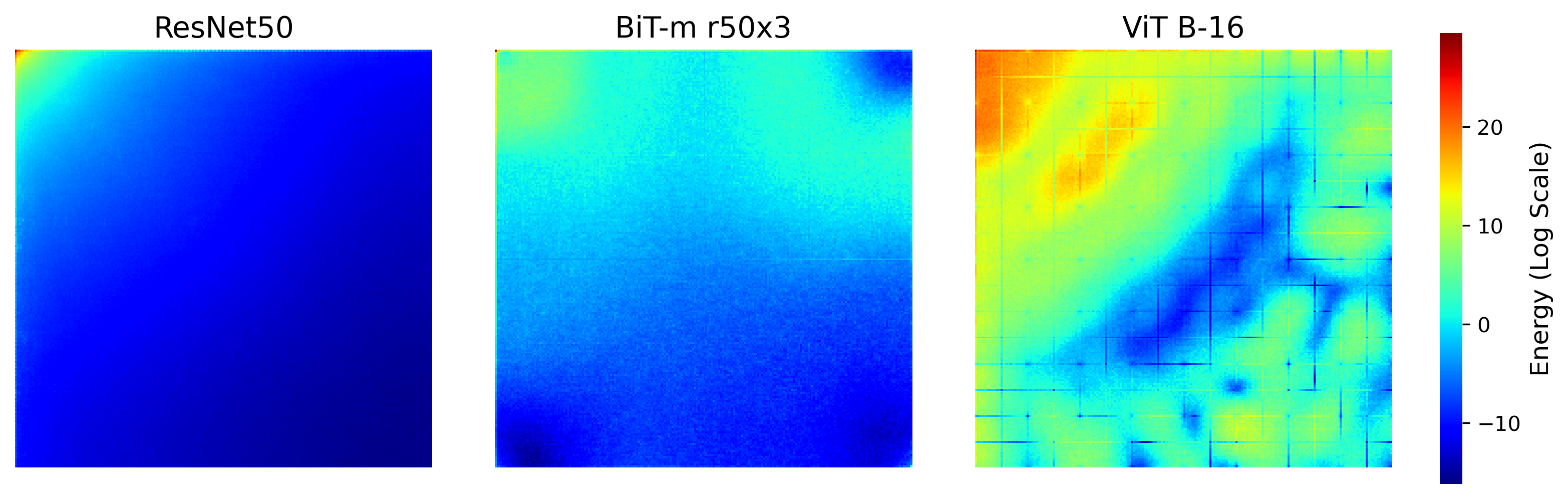} 
\captionof{figure}{Spectral decomposition of adversarial perturbations generated using DeepFool \cite{moosavi2016deepfool}. The top-left/bottom-right quadrants denote low-frequency/high-frequency regions.} 
\label{fig:dct-heatmaps}
\end{minipage}
\vspace{-6mm}
\end{table*}


\subsection{Fourier Spectrum Analysis Reveals Low Sensitivity for ViT}
\label{fourier-analysis}


A common hypothesis about vision models is that they can easily pick up the spurious correlations present inside input data that may be imperceptible and unintuitive to humans 
\cite{jo2017measuring, DBLP:conf/iclr/HendrycksD19}. To measure how ViT holds up with this end of the bargain, we conduct a Fourier analysis \cite{NEURIPS2019_b05b57f6} of ViT, BiT, and our baseline ResNet-50. The experiment goes as follows:

\begin{itemize}[leftmargin=*,noitemsep,topsep=0pt,parsep=0pt,partopsep=0pt]
    \item Generate a Fourier basis vector with varying frequencies. 
    \item Add the basis vector to 1000 randomly sampled images from the ImageNet-1k validation set. 
    \item Record error-rate for every perturbed image and generate a heatmap of the final error matrix.
\end{itemize}

For additional details on this experiment, we refer the reader to \cite{NEURIPS2019_b05b57f6}. In Figure \ref{fig:fourier}, it is noticed that both ViT and BiT stay robust (have low sensitivity) to most of the regions present inside the perturbed images while the baseline ResNet50V2 loses its consistency in the high-frequency regions. The value at location $(i, j)$ shows the error rate on data perturbed by the corresponding Fourier basis noise.

The low sensitivity of ViT and BiT may be attributed to the following factors: (\textbf{a}) Both ViT and BiT are pre-trained on a larger dataset and then fine-tuned on ImageNet-1k. Using a larger dataset during pre-training may be acting as a regularizer here \cite{10.1007/978-3-030-58558-7_29}. (\textbf{b}) Evidence also suggests that increased network width has a positive effect on model robustness \cite{DBLP:conf/iclr/HendrycksD19, hendrycks2021nae}. To get a deeper insight into the heatmaps shown in Figure \ref{fig:fourier}, in Table \ref{tab:percentiles-fourier}, we report error-rate percentiles for the three models under consideration. For a more robust model, we should expect to see lower numbers across all the five different percentiles reported in Table \ref{tab:percentiles-fourier} and we confirm that is indeed the case. This may also help explain the better behavior of BiT and ViT in this experiment.

\subsection{Adversarial Perturbations of ViT Has Wider Spread in Energy Spectrum}

In \cite{NEURIPS2020_1ea97de8}, it is shown that small adversarial perturbations can change the decision boundary of neural networks (especially CNNs) and that adversarial training \cite{madry2017towards} exploits this sensitivity to induce robustness. Furthermore, CNNs primarily exploit discriminative features from the low-frequency regions of the input data. Following \cite{NEURIPS2020_1ea97de8}, we conduct the following experiment on 1000 randomly sampled images from the ImageNet-1k validation set with ResNet-50, BiT-m \texttt{r50x3}, and ViT B-16\footnote{For computational constraints we used smaller BiT and ViT variants for this experiment.}:

\begin{itemize}[leftmargin=*,noitemsep,topsep=0pt,parsep=0pt,partopsep=0pt]
    \item Generate small adversarial perturbations ($\delta$) with DeepFool \cite{moosavi2016deepfool} with a step size of 50\footnote{Rest of the hyperparameters are same as what is specified \textcolor{blue}{https://git.io/JEhpG}.}. 
    \item Change the basis of the perturbations with discrete cosine transform (DCT) to compute the energy spectrum of the perturbations. 
\end{itemize}

This experiment aims to confirm that ViT's perturbations will spread out the whole spectrum, while perturbations of ResNet-50 and BiT will be centered only around the low-frequency regions. This is primarily because ViT has the ability to better exploit information that is only available in a global context. Figure \ref{fig:dct-heatmaps} shows the energy spectrum analysis.
It suggests that to attack ViT, (almost) the entire frequency spectrum needs to be affected, while it is less so for BiT and ResNet-50.

\subsection{ViT Has Smoother Loss Landscape to Input Perturbations}
\label{loss-perturbations}


\begin{figure}[t]
\centering
\vspace{-2mm}
\includegraphics[width=0.7\columnwidth]{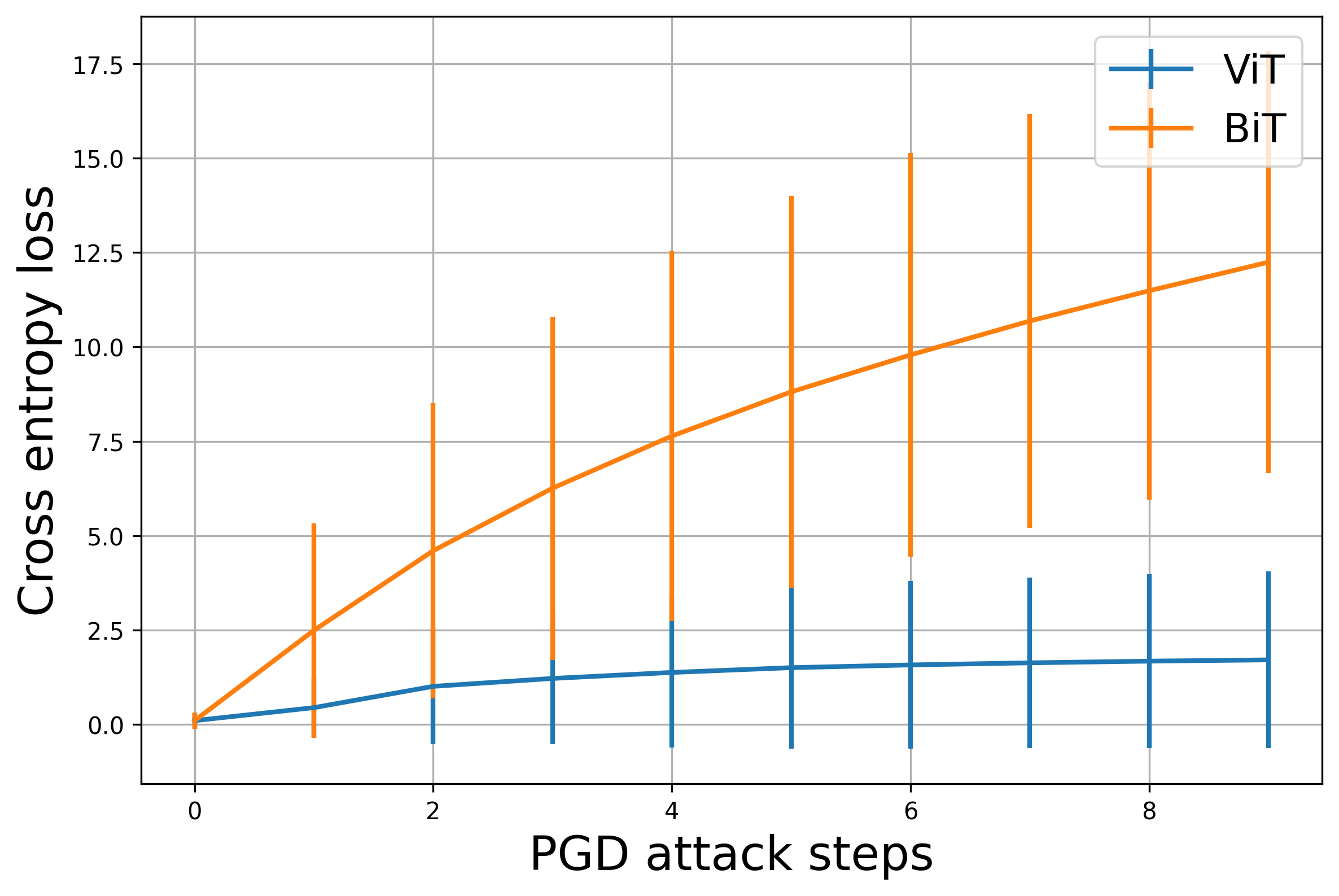} 
\vspace{-2mm}
\caption{Loss progression (mean and standard deviation) ViT (L-16) and BiT-m (\texttt{r101x3}) during PGD attacks \cite{madry2017towards}.}
\label{fig:pgd}
\vspace{-4mm}
\end{figure}

One way to attribute the improved robustness of ViT over BiT is to hypothesize ViT has a smoother loss landscape with respect to input perturbations. Here we explore their loss landscapes based on a common set of 100 ImageNet-1k validation images that are correctly classified by both models. We apply the multi-step projected gradient descent (PGD) attack \cite{madry2017towards} with an $\ell_\infty$ perturbation budget of $\epsilon=0.002$ when normalizing the pixel value range to be between $[-1, 1]$\footnote{We follow the PGD implementation from \textcolor{blue}{https://adversarial-ml-tutorial.org/introduction/}.} (refer to Appendix \ref{perturbations} for details on hyperparameters). Figure \ref{fig:pgd} shows that the classification loss (cross entropy) of ViT increases at a much slower rate than that of BiT as one varies the attack steps, validating our hypothesis of smoother loss landscape to input perturbations.


In summary, in this section, we broadly verify that ViT can yield improved robustness (even with fewer parameters/FLOPS in some cases). This indicates that the use of Transformers can be orthogonal to the known techniques to improve the robustness of vision models \cite{balaji2019instance, NEURIPS2019_32e0bd14, 9156610}.

\section{Conclusion}
Robustness is an important aspect to consider when deploying deep learning models into the wild. This work provides a comprehensive robustness performance assessment of ViTs using 6 different ImageNet datasets and concludes that ViT significantly outperforms its CNN counterpart (BiT) and the baseline ResNet50V2 model. We further conducted 6 new experiments to verify our hypotheses of improved robustness in ViT, including the use of large-scale pre-training and attention module, the ability to recognize randomly masked images, the low sensibility to Fourier spectrum domain perturbation, and the property of wider energy distribution and smoother loss landscape under adversarial input perturbations. Our analyses and findings show novel insights toward understanding the source of robustness and can shed new light on robust neural network architecture design. 

\section*{Acknowledgements}
We are thankful to the Google Developers Experts program\footnote{\textcolor{blue}{https://developers.google.com/programs/experts/}} (specifically Soonson Kwon and Karl Weinmeister) for providing Google Cloud Platform credits to support the experiments. We also thank Justin Gilmer (of Google), Guillermo Ortiz-Jimenez (of EPFL, Switzerland), and Dan Hendrycks (of UC Berkeley) for fruitful discussions. 

\bibliography{adversarial_learning}

\begin{thebibliography}{63}
\providecommand{\natexlab}[1]{#1}

\bibitem[{Abnar and Zuidema(2020)}]{abnar-zuidema-2020-quantifying}
Abnar, S.; and Zuidema, W. 2020.
\newblock Quantifying Attention Flow in Transformers.
\newblock In \emph{Annual Meeting of the Association for Computational
  Linguistics}, 4190--4197.

\bibitem[{Ba, Kiros, and Hinton(2016)}]{ba2016layer}
Ba, J.~L.; Kiros, J.~R.; and Hinton, G.~E. 2016.
\newblock Layer normalization.
\newblock \emph{arXiv preprint arXiv:1607.06450}.

\bibitem[{Bahdanau, Cho, and Bengio(2015)}]{DBLP:journals/corr/BahdanauCB14}
Bahdanau, D.; Cho, K.; and Bengio, Y. 2015.
\newblock Neural Machine Translation by Jointly Learning to Align and
  Translate.
\newblock In \emph{International Conference on Learning Representations}.

\bibitem[{Balaji, Goldstein, and Hoffman(2019)}]{balaji2019instance}
Balaji, Y.; Goldstein, T.; and Hoffman, J. 2019.
\newblock Instance adaptive adversarial training: Improved accuracy tradeoffs
  in neural nets.
\newblock \emph{arXiv preprint arXiv:1910.08051}.

\bibitem[{Bhojanapalli et~al.(2021)Bhojanapalli, Chakrabarti, Glasner, Li,
  Unterthiner, and Veit}]{bhojanapalli2021understanding}
Bhojanapalli, S.; Chakrabarti, A.; Glasner, D.; Li, D.; Unterthiner, T.; and
  Veit, A. 2021.
\newblock Understanding Robustness of Transformers for Image Classification.
\newblock \emph{arXiv preprint arXiv:2103.14586}.

\bibitem[{Cao et~al.(2020)Cao, Xu, Lin, Wei, and Hu}]{cao2020global}
Cao, Y.; Xu, J.; Lin, S.; Wei, F.; and Hu, H. 2020.
\newblock Global Context Networks.
\newblock arXiv:2012.13375.

\bibitem[{Carion et~al.(2020)Carion, Massa, Synnaeve, Usunier, Kirillov, and
  Zagoruyko}]{carion2020endtoend}
Carion, N.; Massa, F.; Synnaeve, G.; Usunier, N.; Kirillov, A.; and Zagoruyko,
  S. 2020.
\newblock End-to-end object detection with transformers.
\newblock In \emph{European Conference on Computer Vision}, 213--229. Springer.

\bibitem[{Carmon et~al.(2019)Carmon, Raghunathan, Schmidt, Duchi, and
  Liang}]{NEURIPS2019_32e0bd14}
Carmon, Y.; Raghunathan, A.; Schmidt, L.; Duchi, J.~C.; and Liang, P.~S. 2019.
\newblock Unlabeled Data Improves Adversarial Robustness.
\newblock In \emph{Advances in Neural Information Processing Systems},
  volume~32.

\bibitem[{Caron et~al.(2021)Caron, Touvron, Misra, J{\'e}gou, Mairal,
  Bojanowski, and Joulin}]{caron2021emerging}
Caron, M.; Touvron, H.; Misra, I.; J{\'e}gou, H.; Mairal, J.; Bojanowski, P.;
  and Joulin, A. 2021.
\newblock Emerging Properties in Self-Supervised Vision Transformers.
\newblock \emph{arXiv preprint arXiv:2104.14294}.

\bibitem[{Chen et~al.(2020)Chen, Radford, Child, Wu, Jun, Luan, and
  Sutskever}]{pmlr-v119-chen20s}
Chen, M.; Radford, A.; Child, R.; Wu, J.; Jun, H.; Luan, D.; and Sutskever, I.
  2020.
\newblock Generative Pretraining From Pixels.
\newblock In \emph{International Conference on Machine Learning}, volume 119,
  1691--1703.

\bibitem[{Chen, Hsieh, and Gong(2021)}]{chen2021vision}
Chen, X.; Hsieh, C.-J.; and Gong, B. 2021.
\newblock When Vision Transformers Outperform ResNets without Pretraining or
  Strong Data Augmentations.
\newblock arXiv:2106.01548.

\bibitem[{Cubuk et~al.(2020)Cubuk, Zoph, Shlens, and Le}]{9150790}
Cubuk, E.~D.; Zoph, B.; Shlens, J.; and Le, Q.~V. 2020.
\newblock Randaugment: Practical automated data augmentation with a reduced
  search space.
\newblock In \emph{IEEE/CVF Conference on Computer Vision and Pattern
  Recognition Workshops}, 3008--3017.

\bibitem[{Deng et~al.(2009)Deng, Dong, Socher, Li, Li, and
  Fei-Fei}]{deng2009imagenet}
Deng, J.; Dong, W.; Socher, R.; Li, L.-J.; Li, K.; and Fei-Fei, L. 2009.
\newblock Imagenet: A large-scale hierarchical image database.
\newblock In \emph{IEEE Conference on Computer Vision and Pattern
  Recognition,}, 248--255.

\bibitem[{Devlin et~al.(2019)Devlin, Chang, Lee, and
  Toutanova}]{devlin-etal-2019-bert}
Devlin, J.; Chang, M.-W.; Lee, K.; and Toutanova, K. 2019.
\newblock {BERT}: Pre-training of Deep Bidirectional Transformers for Language
  Understanding.
\newblock In \emph{Conference of the North {A}merican Chapter of the
  Association for Computational Linguistics: Human Language Technologies},
  4171--4186.

\bibitem[{DeVries and Taylor(2017)}]{devries2017improved}
DeVries, T.; and Taylor, G.~W. 2017.
\newblock Improved regularization of convolutional neural networks with cutout.
\newblock \emph{arXiv preprint arXiv:1708.04552}.

\bibitem[{Dosovitskiy et~al.(2021)Dosovitskiy, Beyer, Kolesnikov, Weissenborn,
  Zhai, Unterthiner, Dehghani, Minderer, Heigold, Gelly, Uszkoreit, and
  Houlsby}]{dosovitskiy2021an}
Dosovitskiy, A.; Beyer, L.; Kolesnikov, A.; Weissenborn, D.; Zhai, X.;
  Unterthiner, T.; Dehghani, M.; Minderer, M.; Heigold, G.; Gelly, S.;
  Uszkoreit, J.; and Houlsby, N. 2021.
\newblock An Image is Worth 16x16 Words: Transformers for Image Recognition at
  Scale.
\newblock In \emph{International Conference on Learning Representations}.

\bibitem[{Foret et~al.(2021)Foret, Kleiner, Mobahi, and
  Neyshabur}]{foret2021sharpnessaware}
Foret, P.; Kleiner, A.; Mobahi, H.; and Neyshabur, B. 2021.
\newblock Sharpness-aware Minimization for Efficiently Improving
  Generalization.
\newblock In \emph{International Conference on Learning Representations}.

\bibitem[{Geirhos et~al.(2019)Geirhos, Rubisch, Michaelis, Bethge, Wichmann,
  and Brendel}]{DBLP:conf/iclr/GeirhosRMBWB19}
Geirhos, R.; Rubisch, P.; Michaelis, C.; Bethge, M.; Wichmann, F.~A.; and
  Brendel, W. 2019.
\newblock ImageNet-trained CNNs are biased towards texture; increasing shape
  bias improves accuracy and robustness.
\newblock In \emph{International Conference on Learning Representations}.

\bibitem[{Hart et~al.(2013)Hart, Schmidt, Klein-Harmeyer, and
  Einh{\"a}user}]{t2013attention}
Hart, B.~M.; Schmidt, H. C. E.~F.; Klein-Harmeyer, I.; and Einh{\"a}user, W.
  2013.
\newblock Attention in natural scenes: contrast affects rapid visual processing
  and fixations alike.
\newblock \emph{Philosophical Transactions of the Royal Society B: Biological
  Sciences}, 368(1628): 20130067.

\bibitem[{{He} et~al.(2016){He}, {Zhang}, {Ren}, and {Sun}}]{7780459}
{He}, K.; {Zhang}, X.; {Ren}, S.; and {Sun}, J. 2016.
\newblock Deep Residual Learning for Image Recognition.
\newblock In \emph{IEEE Conference on Computer Vision and Pattern Recognition},
  770--778.

\bibitem[{He et~al.(2016)He, Zhang, Ren, and
  Sun}]{10.1007/978-3-319-46493-0_38}
He, K.; Zhang, X.; Ren, S.; and Sun, J. 2016.
\newblock Identity Mappings in Deep Residual Networks.
\newblock In Leibe, B.; Matas, J.; Sebe, N.; and Welling, M., eds.,
  \emph{European Conference on Computer Vision}, 630--645. Springer.

\bibitem[{Hendrycks et~al.(2020)Hendrycks, Basart, Mu, Kadavath, Wang, Dorundo,
  Desai, Zhu, Parajuli, Guo, Song, Steinhardt, and Gilmer}]{hendrycks2020many}
Hendrycks, D.; Basart, S.; Mu, N.; Kadavath, S.; Wang, F.; Dorundo, E.; Desai,
  R.; Zhu, T.; Parajuli, S.; Guo, M.; Song, D.; Steinhardt, J.; and Gilmer, J.
  2020.
\newblock The Many Faces of Robustness: A Critical Analysis of
  Out-of-Distribution Generalization.
\newblock \emph{arXiv preprint arXiv:2006.16241}.

\bibitem[{Hendrycks and Dietterich(2019)}]{DBLP:conf/iclr/HendrycksD19}
Hendrycks, D.; and Dietterich, T.~G. 2019.
\newblock Benchmarking Neural Network Robustness to Common Corruptions and
  Perturbations.
\newblock In \emph{International Conference on Learning Representations}.

\bibitem[{Hendrycks and Gimpel(2016)}]{hendrycks2020gaussian}
Hendrycks, D.; and Gimpel, K. 2016.
\newblock Gaussian error linear units (gelus).
\newblock \emph{arXiv preprint arXiv:1606.08415}.

\bibitem[{Hendrycks* et~al.(2020)Hendrycks*, Mu*, Cubuk, Zoph, Gilmer, and
  Lakshminarayanan}]{hendrycks*2020augmix}
Hendrycks*, D.; Mu*, N.; Cubuk, E.~D.; Zoph, B.; Gilmer, J.; and
  Lakshminarayanan, B. 2020.
\newblock AugMix: A Simple Method to Improve Robustness and Uncertainty under
  Data Shift.
\newblock In \emph{International Conference on Learning Representations}.

\bibitem[{Hendrycks et~al.(2021)Hendrycks, Zhao, Basart, Steinhardt, and
  Song}]{hendrycks2021nae}
Hendrycks, D.; Zhao, K.; Basart, S.; Steinhardt, J.; and Song, D. 2021.
\newblock Natural Adversarial Examples.
\newblock \emph{Conference on Computer Vision and Pattern Recognition}.

\bibitem[{Hinton, Vinyals, and Dean(2015)}]{44873}
Hinton, G.; Vinyals, O.; and Dean, J. 2015.
\newblock Distilling the Knowledge in a Neural Network.
\newblock In \emph{NeurIPS Deep Learning and Representation Learning Workshop}.

\bibitem[{Hu et~al.(2018)Hu, Shen, Albanie, Sun, and
  Vedaldi}]{NEURIPS2018_dc363817}
Hu, J.; Shen, L.; Albanie, S.; Sun, G.; and Vedaldi, A. 2018.
\newblock Gather-Excite: Exploiting Feature Context in Convolutional Neural
  Networks.
\newblock In Bengio, S.; Wallach, H.; Larochelle, H.; Grauman, K.;
  Cesa-Bianchi, N.; and Garnett, R., eds., \emph{Advances in Neural Information
  Processing Systems}, volume~31. Curran Associates, Inc.

\bibitem[{Huang et~al.(2016)Huang, Sun, Liu, Sedra, and
  Weinberger}]{huang2016deep}
Huang, G.; Sun, Y.; Liu, Z.; Sedra, D.; and Weinberger, K.~Q. 2016.
\newblock Deep networks with stochastic depth.
\newblock In \emph{European conference on computer vision}, 646--661. Springer.

\bibitem[{Ioffe and Szegedy(2015)}]{pmlr-v37-ioffe15}
Ioffe, S.; and Szegedy, C. 2015.
\newblock Batch Normalization: Accelerating Deep Network Training by Reducing
  Internal Covariate Shift.
\newblock In \emph{International Conference on Machine Learning}, volume~37,
  448--456.

\bibitem[{Jiang et~al.(2021)Jiang, Hou, Yuan, Zhou, Jin, Wang, and
  Feng}]{jiang2021token}
Jiang, Z.; Hou, Q.; Yuan, L.; Zhou, D.; Jin, X.; Wang, A.; and Feng, J. 2021.
\newblock Token labeling: Training a 85.5\% top-1 accuracy vision transformer
  with 56m parameters on imagenet.
\newblock \emph{arXiv preprint arXiv:2104.10858}.

\bibitem[{Jo and Bengio(2017)}]{jo2017measuring}
Jo, J.; and Bengio, Y. 2017.
\newblock Measuring the tendency of cnns to learn surface statistical
  regularities.
\newblock \emph{arXiv preprint arXiv:1711.11561}.

\bibitem[{Kingma and Ba(2015)}]{kingma2014adam}
Kingma, D.; and Ba, J. 2015.
\newblock Adam: A method for stochastic optimization.
\newblock \emph{International Conference on Learning Representations}.

\bibitem[{Kolesnikov et~al.(2020)Kolesnikov, Beyer, Zhai, Puigcerver, Yung,
  Gelly, and Houlsby}]{10.1007/978-3-030-58558-7_29}
Kolesnikov, A.; Beyer, L.; Zhai, X.; Puigcerver, J.; Yung, J.; Gelly, S.; and
  Houlsby, N. 2020.
\newblock Big Transfer (BiT): General Visual Representation Learning.
\newblock In \emph{European Conference on Computer Vision}, 491--507.

\bibitem[{Krizhevsky, Sutskever, and Hinton(2012)}]{NIPS2012_c399862d}
Krizhevsky, A.; Sutskever, I.; and Hinton, G.~E. 2012.
\newblock ImageNet Classification with Deep Convolutional Neural Networks.
\newblock In \emph{Advances in Neural Information Processing Systems},
  volume~25.

\bibitem[{Li et~al.(2019)Li, Wang, Hu, and Yang}]{8954149}
Li, X.; Wang, W.; Hu, X.; and Yang, J. 2019.
\newblock Selective Kernel Networks.
\newblock In \emph{2019 IEEE/CVF Conference on Computer Vision and Pattern
  Recognition (CVPR)}, 510--519.

\bibitem[{Liu et~al.(2021)Liu, Lin, Cao, Hu, Wei, Zhang, Lin, and
  Guo}]{liu2021swin}
Liu, Z.; Lin, Y.; Cao, Y.; Hu, H.; Wei, Y.; Zhang, Z.; Lin, S.; and Guo, B.
  2021.
\newblock Swin Transformer: Hierarchical Vision Transformer using Shifted
  Windows.
\newblock arXiv:2103.14030.

\bibitem[{Madry et~al.(2018)Madry, Makelov, Schmidt, Tsipras, and
  Vladu}]{madry2017towards}
Madry, A.; Makelov, A.; Schmidt, L.; Tsipras, D.; and Vladu, A. 2018.
\newblock Towards Deep Learning Models Resistant to Adversarial Attacks.
\newblock \emph{International Conference on Learning Representations}.

\bibitem[{Mahmood, Mahmood, and Van~Dijk(2021)}]{mahmood2021robustness}
Mahmood, K.; Mahmood, R.; and Van~Dijk, M. 2021.
\newblock On the Robustness of Vision Transformers to Adversarial Examples.
\newblock \emph{arXiv preprint arXiv:2104.02610}.

\bibitem[{Moosavi-Dezfooli, Fawzi, and Frossard(2016)}]{moosavi2016deepfool}
Moosavi-Dezfooli, S.-M.; Fawzi, A.; and Frossard, P. 2016.
\newblock Deepfool: a simple and accurate method to fool deep neural networks.
\newblock In \emph{IEEE Conference on Computer Vision and Pattern Recognition},
  2574--2582.

\bibitem[{Ortiz-Jimenez et~al.(2020)Ortiz-Jimenez, Modas, Moosavi, and
  Frossard}]{NEURIPS2020_1ea97de8}
Ortiz-Jimenez, G.; Modas, A.; Moosavi, S.-M.; and Frossard, P. 2020.
\newblock Hold me tight! Influence of discriminative features on deep network
  boundaries.
\newblock In \emph{Advances in Neural Information Processing Systems},
  volume~33, 2935--2946.

\bibitem[{Parmar et~al.(2018)Parmar, Vaswani, Uszkoreit, Kaiser, Shazeer, Ku,
  and Tran}]{pmlr-v80-parmar18a}
Parmar, N.; Vaswani, A.; Uszkoreit, J.; Kaiser, L.; Shazeer, N.; Ku, A.; and
  Tran, D. 2018.
\newblock Image Transformer.
\newblock In \emph{International Conference on Machine Learning}, volume~80,
  4055--4064.

\bibitem[{Qiao et~al.(2019)Qiao, Wang, Liu, Shen, and
  Yuille}]{qiao2020microbatch}
Qiao, S.; Wang, H.; Liu, C.; Shen, W.; and Yuille, A. 2019.
\newblock Micro-Batch Training with Batch-Channel Normalization and Weight
  Standardization.
\newblock \emph{arXiv preprint arXiv:1903.10520}.

\bibitem[{Radford et~al.(2021)Radford, Kim, Hallacy, Ramesh, Goh, Agarwal,
  Sastry, Askell, Mishkin, Clark, Krueger, and Sutskever}]{radford2021learning}
Radford, A.; Kim, J.~W.; Hallacy, C.; Ramesh, A.; Goh, G.; Agarwal, S.; Sastry,
  G.; Askell, A.; Mishkin, P.; Clark, J.; Krueger, G.; and Sutskever, I. 2021.
\newblock Learning Transferable Visual Models From Natural Language
  Supervision.
\newblock arXiv:2103.00020.

\bibitem[{Radosavovic et~al.(2020)Radosavovic, Kosaraju, Girshick, He, and
  Dollar}]{9156494}
Radosavovic, I.; Kosaraju, R.; Girshick, R.; He, K.; and Dollar, P. 2020.
\newblock Designing Network Design Spaces.
\newblock In \emph{IEEE Conference on Computer Vision and Pattern Recognition},
  10425--10433.

\bibitem[{Raghu et~al.(2021)Raghu, Unterthiner, Kornblith, Zhang, and
  Dosovitskiy}]{raghu2021vision}
Raghu, M.; Unterthiner, T.; Kornblith, S.; Zhang, C.; and Dosovitskiy, A. 2021.
\newblock Do Vision Transformers See Like Convolutional Neural Networks?
\newblock arXiv:2108.08810.

\bibitem[{Russakovsky et~al.(2015)Russakovsky, Deng, Su, Krause, Satheesh, Ma,
  Huang, Karpathy, Khosla, Bernstein et~al.}]{russakovsky2015imagenet}
Russakovsky, O.; Deng, J.; Su, H.; Krause, J.; Satheesh, S.; Ma, S.; Huang, Z.;
  Karpathy, A.; Khosla, A.; Bernstein, M.; et~al. 2015.
\newblock Imagenet large scale visual recognition challenge.
\newblock \emph{International Journal of Computer Vision}, 115(3): 211--252.

\bibitem[{Selvaraju et~al.(2017)Selvaraju, Cogswell, Das, Vedantam, Parikh, and
  Batra}]{8237336}
Selvaraju, R.~R.; Cogswell, M.; Das, A.; Vedantam, R.; Parikh, D.; and Batra,
  D. 2017.
\newblock Grad-CAM: Visual Explanations from Deep Networks via Gradient-Based
  Localization.
\newblock In \emph{IEEE International Conference on Computer Vision}, 618--626.

\bibitem[{Shao et~al.(2021)Shao, Shi, Yi, Chen, and
  Hsieh}]{shao2021adversarial}
Shao, R.; Shi, Z.; Yi, J.; Chen, P.-Y.; and Hsieh, C.-J. 2021.
\newblock On the Adversarial Robustness of Visual Transformers.
\newblock \emph{arXiv preprint arXiv:2103.15670}.

\bibitem[{Srivastava et~al.(2014)Srivastava, Hinton, Krizhevsky, Sutskever, and
  Salakhutdinov}]{JMLR:v15:srivastava14a}
Srivastava, N.; Hinton, G.; Krizhevsky, A.; Sutskever, I.; and Salakhutdinov,
  R. 2014.
\newblock Dropout: A Simple Way to Prevent Neural Networks from Overfitting.
\newblock \emph{Journal of Machine Learning Research}, 15(56): 1929--1958.

\bibitem[{Steiner et~al.(2021)Steiner, Kolesnikov, Zhai, Wightman, Uszkoreit,
  and Beyer}]{steiner2021train}
Steiner, A.; Kolesnikov, A.; Zhai, X.; Wightman, R.; Uszkoreit, J.; and Beyer,
  L. 2021.
\newblock How to train your ViT? Data, Augmentation, and Regularization in
  Vision Transformers.
\newblock arXiv:2106.10270.

\bibitem[{{Sun} et~al.(2017){Sun}, {Shrivastava}, {Singh}, and
  {Gupta}}]{8237359}
{Sun}, C.; {Shrivastava}, A.; {Singh}, S.; and {Gupta}, A. 2017.
\newblock Revisiting Unreasonable Effectiveness of Data in Deep Learning Era.
\newblock In \emph{IEEE International Conference on Computer Vision}, 843--852.

\bibitem[{Tan and Le(2021)}]{pmlr-v139-tan21a}
Tan, M.; and Le, Q. 2021.
\newblock EfficientNetV2: Smaller Models and Faster Training.
\newblock In Meila, M.; and Zhang, T., eds., \emph{Proceedings of the 38th
  International Conference on Machine Learning}, volume 139 of
  \emph{Proceedings of Machine Learning Research}, 10096--10106. PMLR.

\bibitem[{Touvron et~al.(2020)Touvron, Cord, Douze, Massa, Sablayrolles, and
  J{\'e}gou}]{touvron2021training}
Touvron, H.; Cord, M.; Douze, M.; Massa, F.; Sablayrolles, A.; and J{\'e}gou,
  H. 2020.
\newblock Training data-efficient image transformers \& distillation through
  attention.
\newblock \emph{arXiv preprint arXiv:2012.12877}.

\bibitem[{Trinh, Luong, and Le(2019)}]{trinh2019selfie}
Trinh, T.~H.; Luong, M.-T.; and Le, Q.~V. 2019.
\newblock Selfie: Self-supervised pretraining for image embedding.
\newblock \emph{arXiv preprint arXiv:1906.02940}.

\bibitem[{Tuli et~al.(2021)Tuli, Dasgupta, Grant, and
  Griffiths}]{tuli2021convolutional}
Tuli, S.; Dasgupta, I.; Grant, E.; and Griffiths, T.~L. 2021.
\newblock Are Convolutional Neural Networks or Transformers more like human
  vision?
\newblock Accepted at CogSci 2021.

\bibitem[{Vaswani et~al.(2017)Vaswani, Shazeer, Parmar, Uszkoreit, Jones,
  Gomez, Kaiser, and Polosukhin}]{NIPS2017_3f5ee243}
Vaswani, A.; Shazeer, N.; Parmar, N.; Uszkoreit, J.; Jones, L.; Gomez, A.~N.;
  Kaiser, L.~u.; and Polosukhin, I. 2017.
\newblock Attention is All you Need.
\newblock In \emph{Advances in Neural Information Processing Systems},
  volume~30.

\bibitem[{Wu and He(2018)}]{Wu_2018_ECCV}
Wu, Y.; and He, K. 2018.
\newblock Group Normalization.
\newblock In \emph{European Conference on Computer Vision}, 3--19.

\bibitem[{Xiao et~al.(2021)Xiao, Engstrom, Ilyas, and Madry}]{xiao2020noise}
Xiao, K.; Engstrom, L.; Ilyas, A.; and Madry, A. 2021.
\newblock Noise or Signal: The Role of Image Backgrounds in Object Recognition.
\newblock \emph{International Conference on Learning Representations}.

\bibitem[{Xie et~al.(2020)Xie, Luong, Hovy, and Le}]{9156610}
Xie, Q.; Luong, M.-T.; Hovy, E.; and Le, Q.~V. 2020.
\newblock Self-Training With Noisy Student Improves ImageNet Classification.
\newblock In \emph{IEEE Conference on Computer Vision and Pattern Recognition},
  10684--10695.

\bibitem[{Yin et~al.(2019)Yin, Gontijo~Lopes, Shlens, Cubuk, and
  Gilmer}]{NEURIPS2019_b05b57f6}
Yin, D.; Gontijo~Lopes, R.; Shlens, J.; Cubuk, E.~D.; and Gilmer, J. 2019.
\newblock A {Fourier} Perspective on Model Robustness in Computer Vision.
\newblock In \emph{Advances in Neural Information Processing Systems},
  volume~32.

\bibitem[{Yuan et~al.(2021)Yuan, Chen, Wang, Yu, Shi, Tay, Feng, and
  Yan}]{yuan2021tokenstotoken}
Yuan, L.; Chen, Y.; Wang, T.; Yu, W.; Shi, Y.; Tay, F.~E.; Feng, J.; and Yan,
  S. 2021.
\newblock Tokens-to-token vit: Training vision transformers from scratch on
  imagenet.
\newblock \emph{arXiv preprint arXiv:2101.11986}.

\bibitem[{Zagoruyko and Komodakis(2016)}]{zagoruyko2017wide}
Zagoruyko, S.; and Komodakis, N. 2016.
\newblock Wide residual networks.
\newblock \emph{arXiv preprint arXiv:1605.07146}.

\end{thebibliography}

\clearpage

\appendix

\section*{Appendix}

\section{ViT Preliminaries}
\label{sec_prelim}
\paragraph{Multi-head Self Attention (MHSA).}

Central to ViT's model design is self-attention \cite{DBLP:journals/corr/BahdanauCB14}. Here, we first compute three quantities from linear projections ($X \in \mathbb{R}^{N \times D}$): (\textbf{i}) \textbf{Q}uery = $X W_{\mathrm{Q}}$,  (\textbf{ii}) \textbf{K}ey = $X W_{\mathrm{K}}$ , and (\textbf{iii}) \textbf{V}alue = $X W_{\mathrm{V}}$, where $W_{\mathrm{Q}}$, $W_{\mathrm{K}}$, and $W_{\mathrm{V}}$ are linear transformations. The linear projections ($X$) are computed from batches of the original input data. Self-attention takes these three input quantities and returns an output matrix ($N \times d$) weighted by attention scores using
\eqref{attn-eq}:
\begin{equation}
\label{attn-eq}
\operatorname{Attention}(Q, K, V)=\operatorname{Softmax}\left(Q K^{\top} / \sqrt{d}\right) V
\end{equation}
This form of attention is also popularly referred to as the "scaled dot-product attention" \cite{NIPS2017_3f5ee243}. One important aspect of self-attention is that it operates between all pairs of elements within an input. In summary, a single attention layer tries to find out how to best align the keys to the queries and quantifies this finding in the form of attention scores. These scores are then multiplied with the values to obtain the final output. To enable feature-rich hierarchical learning, $h$ self-attention layers (or so-called "heads") are stacked together producing an output of $N \times d h$. This output is then fed through a linear transformation layer that produces the final output of $N \times d$ from MHSA. MHSA then forms the core Transformer block. 

\paragraph{Transformer block.} A single transformer block is composed of MHSA, Layer Normalization (LN) \cite{ba2016layer}, feed-forward network (FFN), and skip connections \cite{7780459}. It is implemented using (\ref{transformer-block}):
\begin{equation}
\label{transformer-block}
\begin{aligned}
\mathbf{z}_{\ell}^{\prime} &=\operatorname{MHSA}\left(\operatorname{LN}\left(\mathbf{z}_{\ell-1}\right)\right)+\mathbf{z}_{\ell-1} ;~
\\
\mathbf{z}_{\ell} &=\operatorname{FFN}\left(\operatorname{LN}\left(\mathbf{z}_{\ell}^{\prime}\right)\right)+\mathbf{z}_{\ell}^{\prime} ;~
\mathbf{y} =\operatorname{LN}\left(\mathbf{z}_{L}^{0}\right),
\end{aligned}
\end{equation}
where $\ell \in \{0,1,\ldots,L\}$ is the layer index and $L$ is the number of hidden layers.

The FFN is comprised of two linear layers with a GELU non-linearity \cite{hendrycks2020gaussian} in between them. We refer the reader to Figure 1 of \cite{dosovitskiy2021an} for a pictorial overview of the Transformer block. Next, we discuss the class-token (learned version of which is represented as ${z}_{L}^{0}$ in \eqref{transformer-block}) and how images are fed to a Transformer block with patch encoding. 

\paragraph{Class token and encoded patches of images.}

Inspired by BERT \cite{devlin-etal-2019-bert}, a class token is prepended to the image patches and it flows through the entirety of ViT. It is initialized as ${z}_{0}^{0}$ and serves as the final representation of the image patches which is then passed to task head.
Transformers can only process sequences of inputs. Consider an image of $N \times N$ shape. If we were to extract patches of shape $P \times P$ then the total number of patches would be $(N / P)^2$ (see Appendix \ref{apdnx:patches} for more details). 

A Transformer block processes these patches in parallel which makes it invariant to the order in which the patches would appear. Since locality is not just desirable but also is necessary especially in images, a learnable position encoder is used to get linear projections of the positions of the image patches. These projections are combined with the linear projections of the actual image patches and are then fed to the subsequent Transformer blocks. In \cite{dosovitskiy2021an}, the authors also investigate hybrid models wherein the patch encoding is applied on feature maps computed using a CNN. However, in this work, we do not consider those.

\section{Image Patches}
\label{apdnx:patches}

\begin{figure}[ht]
\centering
\includegraphics[width=0.8\columnwidth]{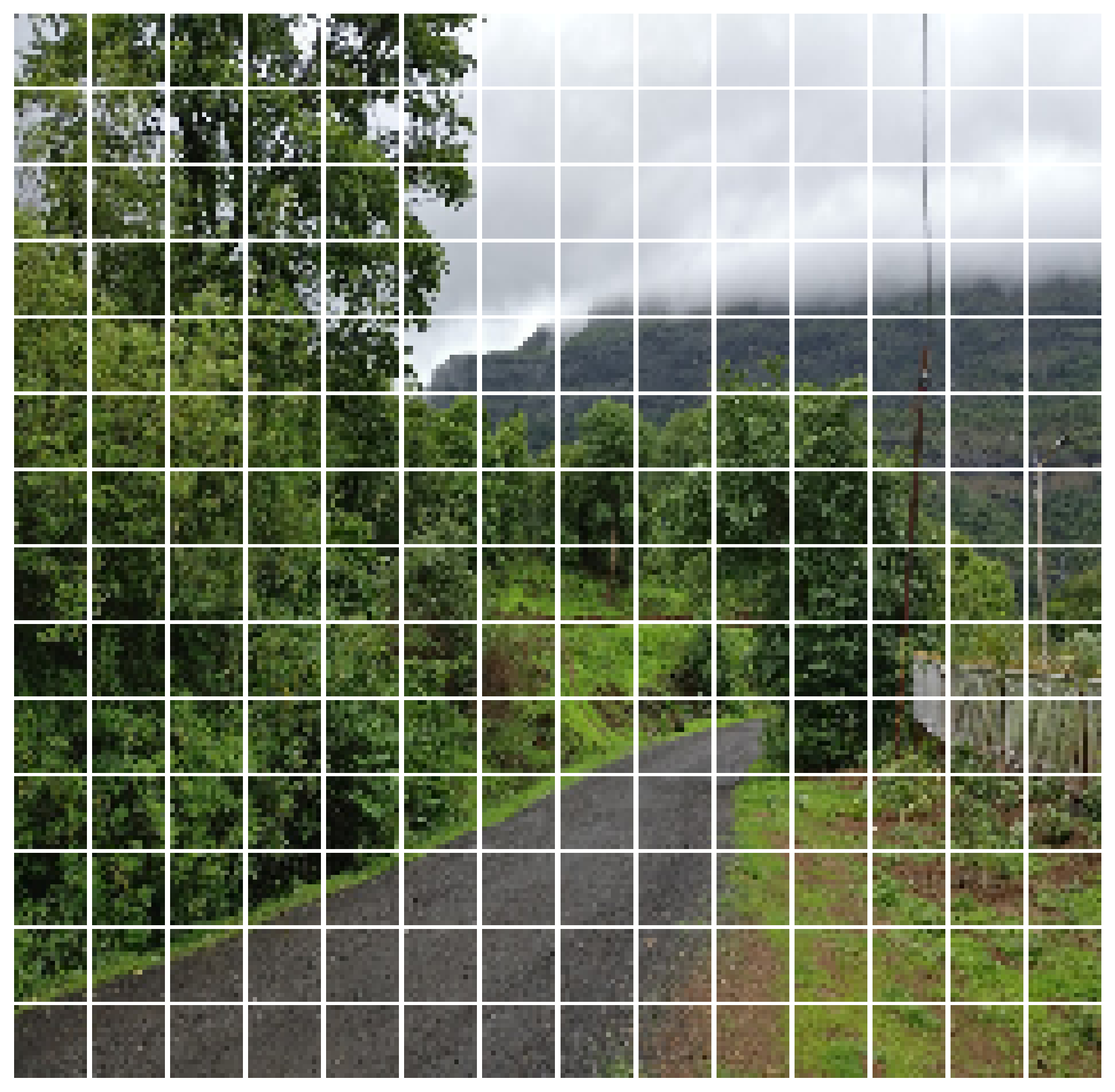} 
\caption{A sample image divided into patches.}
\label{fig:image-patches}
\end{figure}

In ViT, the input images are divided into small patches as depicted in Figure \ref{fig:image-patches}. Here, the original image is of 224 $\times$ 224 shape and each patch is of $16 \times 16$ shape. This gives us a total of 196 patches. Since we are dealing with RGB images here, we also need to consider the channel dimension. So, in total, these $16 \times 16$ patches (with 3 channels) are flattened into a dimension of 768 ($16 \times 16 \times 3$) using a linear transformation. The spatial information of the patches gets lost due to this and to mitigate that position encoding is used. For visual depictions of how the patch encodings operate with each other, we refer the reader to the Figure 7 of \cite{dosovitskiy2021an}. 

\section{Additional Results on Attention}
\label{appen_attention}

\begin{figure*}[t]
\begin{minipage}[t]{.45\textwidth}
\centering
\includegraphics[width=1\columnwidth]{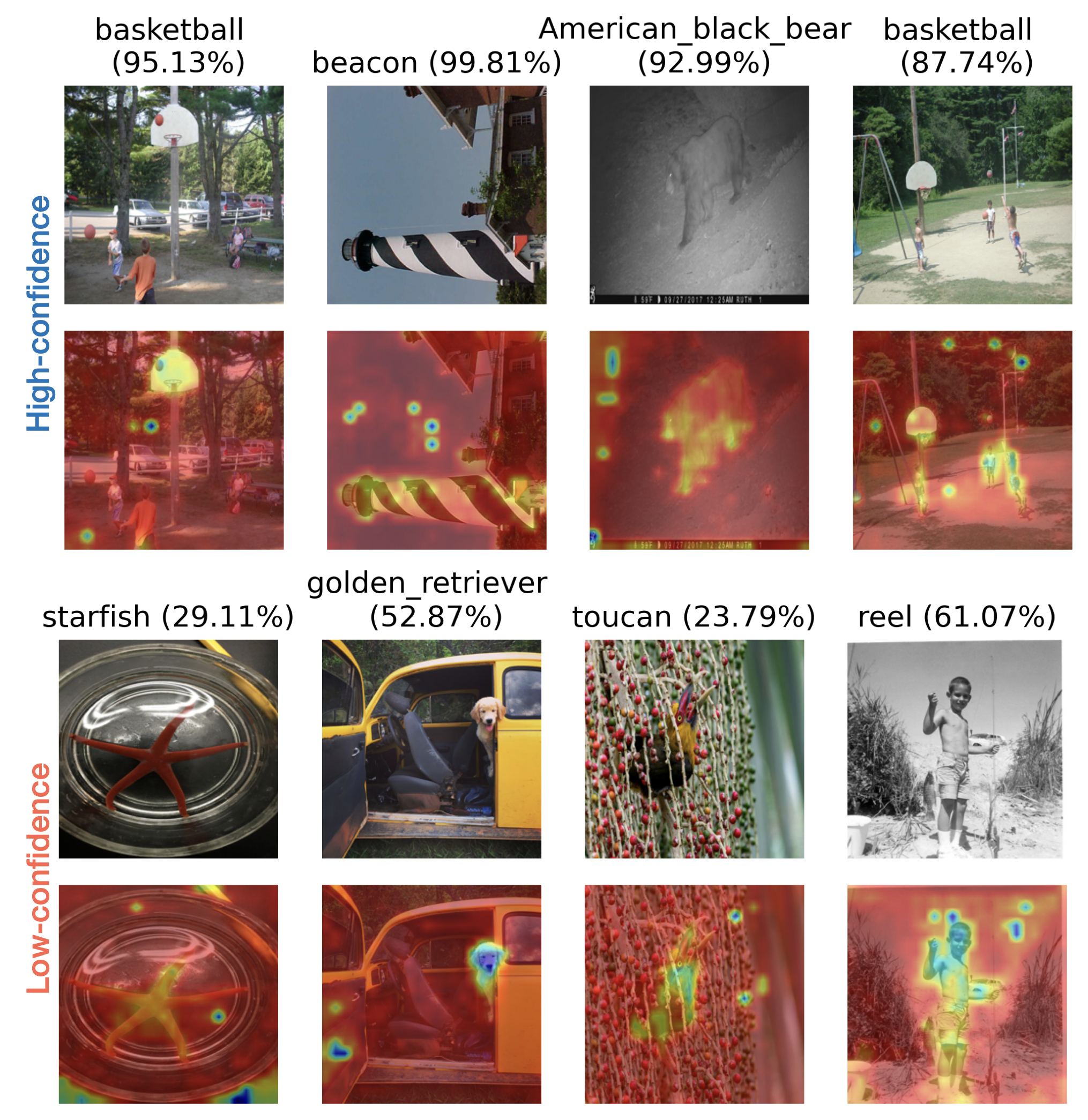} 
\caption{Visualization of the attention maps of ViT on images (top rows) from ImageNet-A. 
} 
\label{fig:attn-map-nae}
\end{minipage}
\hfill
\begin{minipage}[t]{.45\textwidth}
\centering
\includegraphics[width=0.97\columnwidth]{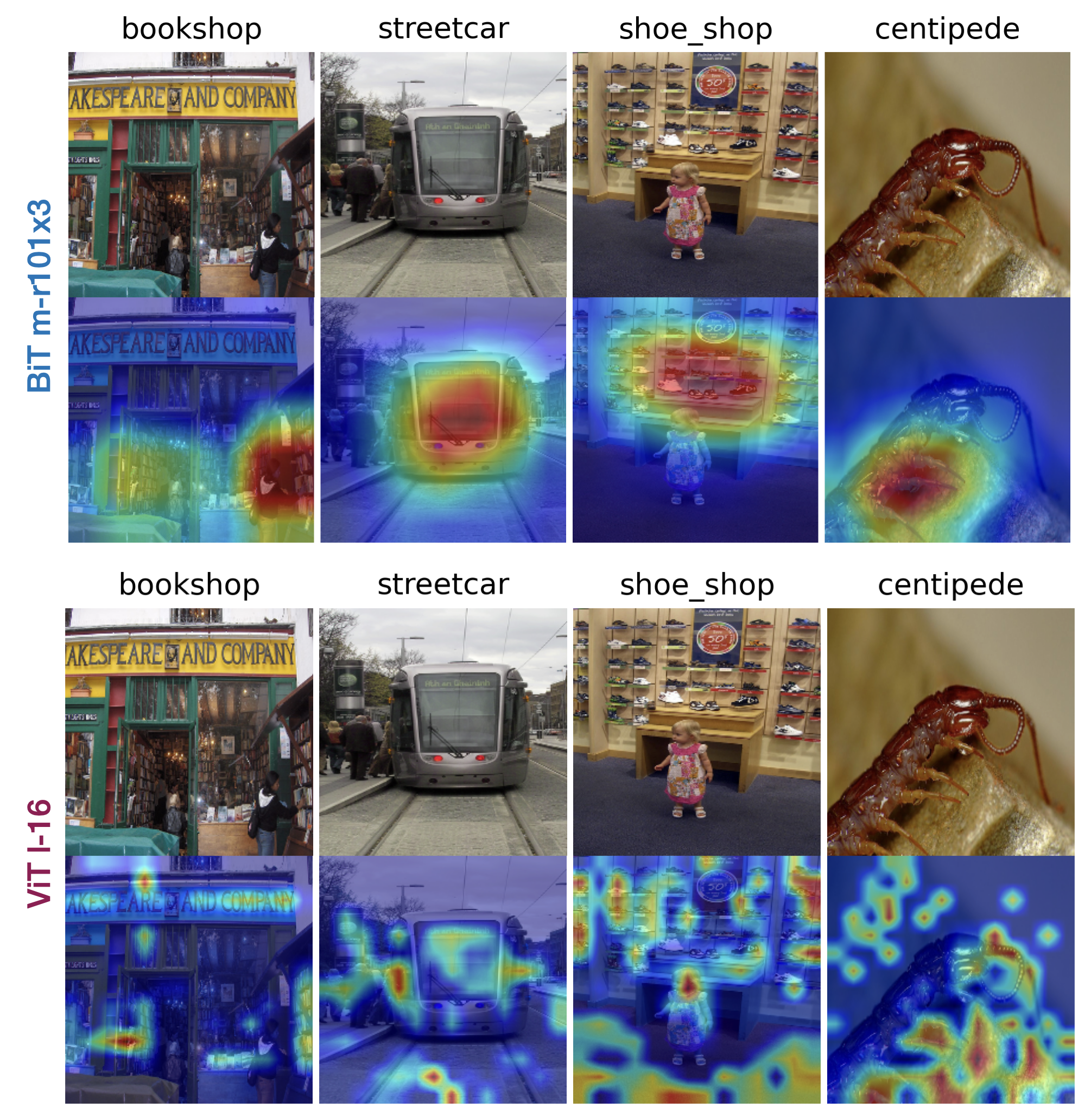} 
\caption{Grad-CAM results for the images where both BiT and ViT give correct predictions. 
}
\label{fig:gradcam}
\end{minipage}
\vspace{-4mm}
\end{figure*}

In this section we present additional experiments showing connections between attention and robustness. In particular, we make use of Attention Rollout \cite{abnar-zuidema-2020-quantifying} to visualize the attention maps for two different cases: (\textbf{a}) where ViT yields high-confidence \textit{correct} predictions  and (\textbf{b}) where ViT yields low-confidence \textit{correct} predictions. Figure \ref{fig:attn-map-nae}\footnote{For this study, we used the ViT L-16 variant as it yields the best performance on the ImageNet-A dataset (refer to Figure \ref{fig:imagenet-a}).} presents a few visualizations of this study. 

It is surprising to see that even under such dark lighting conditions, ViT is able to make the correct predictions for the "American Black Beer" class (second-last plot from Figure \ref{fig:attn-map-nae} (top)). On the other hand, for the low-confidence cases, although ViT is still able to produce the correct predictions it is not very clear where it is putting its focus. For example, consider the last plot from Figure \ref{fig:attn-map-nae} (bottom). ViT draws all its attention to the individual that is standing and \textit{not} on the reel they are holding. 

To further investigate the representations learned by ViT and to better understand the spread of the attention span of ViT, we apply Grad-CAM \cite{8237336} and compare the results to that of BiT. For BiT, we take the last convolutional block for computing the gradients with respect to the target class. Our comparative results are presented in Figure \ref{fig:gradcam}. But we cannot apply these steps directly to ViT because of its structure. Hence we compute the class activation gradients of the last attention block reshaped to fit the computations of Grad-CAM\footnote{For this, we follow the implementation of the \texttt{pytorch\_grad\_cam} library.}. 


From Figure \ref{fig:gradcam}, it can be noticed that ViT tries to maintain a global context in order to generate the predictions while the explanations for BiT are more local and central. For example, consider the image predicted as "bookshop" in Figure \ref{fig:gradcam}. We can observe that ViT uses information from different parts of the image to determine the target class. Objects of interest inside images may not be always centrally aligned with the respect of the objects of images. Besides, capturing long-range dependencies is desirable when dealing with tasks like object detection and segmentation \cite{carion2020endtoend}. This is why we hypothesize that ViT should be able to perform well even when some seemingly attentive regions of an image are masked out. We study this hypothesis in Section \ref{masking-cutout}. 
It should also be noted that there are some spurious attention regions that are not very explanatory (refer to the image predicted as "centipede") and may lead to future research\footnote{DINO \cite{caron2021emerging} shows when ViT is trained with self-supervised objective on a larger data corpus, its self-attention span is made even more pronounced than their self-supervised counterparts. With the virtue of self-supervision, DINO is able to perfectly segment objects of interest from an image \textit{without} any supervision.}. 

\section{Additional Experiments for Pre-training}
\label{appen_pre-training}
To study the effects of better pre-training on ViT and BiT even further, we conduct the following experiment:

\begin{itemize}[leftmargin=*]
    \item We take all the images from the ImageNet-A dataset \cite{hendrycks2021nae}. This dataset is chosen because it has many properties that are desirable for testing a model's robustness capabilities: (\textbf{a}) Objects of interest present in many of the images in the dataset are not centrally oriented, (\textbf{b}) Multiple images have multiple objects present inside them which makes it harder for a model to associate the images with discrete individual categories, (\textbf{c}) Different images have varying amount of textures that can act as spurious correlations for neural nets to produce misclassifications \cite{DBLP:conf/iclr/GeirhosRMBWB19}, and (\textbf{d}) These traits are not very uncommon to catch in a large portion of real-world images.  
    
    \item Run the images through different variants of BiT and ViT and record the top-1 accuracies. For this experiment, we also include the \texttt{BiT-s} variants that are pre-trained on the ImageNet-1k dataset. 
\end{itemize}

Our findings for the above-described experiment are summarized in Table \ref{tab:pre-training-nae}. 
When a longer pre-training schedule is coupled with a larger pre-training dataset it can be helpful in improving a model's performance on the ImageNet-A dataset. Another noticeable trend is that as the model capacity increases the performance also improves. Among all the variants of the different models we present in Table To study the effects of better pre-training on ViT and BiT even further , we conduct the following experiment:

\begin{table}[h]
\centering
\adjustbox{max width=1\columnwidth}{
\begin{tabular}{@{}cccc@{}}
\toprule
\textbf{BiT-s Variant} & \textbf{\begin{tabular}[c]{@{}c@{}}BiT-s Top-1\\ Acc (\%)\end{tabular}} & \textbf{BiT-m $-$ BiT-s} & \textbf{ViT $-$ BiT-m} \\ \midrule
\multicolumn{1}{c}{s-r50x1}  & \multicolumn{1}{c}{2.6}  & \multicolumn{1}{c}{+1.57} & \multicolumn{1}{c}{+22.5 (ViT B-16)}  \\ \midrule
\multicolumn{1}{c}{s-r50x3}  & \multicolumn{1}{c}{3.2}  & \multicolumn{1}{c}{+5.43} & \multicolumn{1}{c}{+7.63 (ViT B-32)}  \\ \midrule
\multicolumn{1}{c}{s-r101x1} & \multicolumn{1}{c}{3.11} & \multicolumn{1}{c}{+3.3}  & \multicolumn{1}{c}{+21.69 (ViT L-16)} \\ \midrule
\multicolumn{1}{c}{s-r101x3} & \multicolumn{1}{c}{4.29} & \multicolumn{1}{c}{+6.14} & \multicolumn{1}{c}{+8.62 (ViT L-32)}  \\ \midrule
 s-r152x4                       & 4.64                      & +8.51                      & NA                                    \\ \bottomrule
\end{tabular}
}
\vspace{-2mm}
\caption{Relative improvements as achieved by \texttt{BiT-m} and ViT variants. BiT \texttt{m-r101x3} is comparable to ViT L-16, and ViT L-32 with respect to the number of model parameters. }
\label{tab:pre-training-nae}
\end{table}

\section{Additional Examples with Grad-CAM}
\label{gradcam-more}

To make our arguments in Appendix \ref{appen_attention} more concrete, we provide additional results from Grad-CAM in Figure \ref{gradcam-more}. All the original images are from ImageNet-1k validation set.

\begin{figure}[ht]
\centering
\includegraphics[width=1.0\columnwidth]{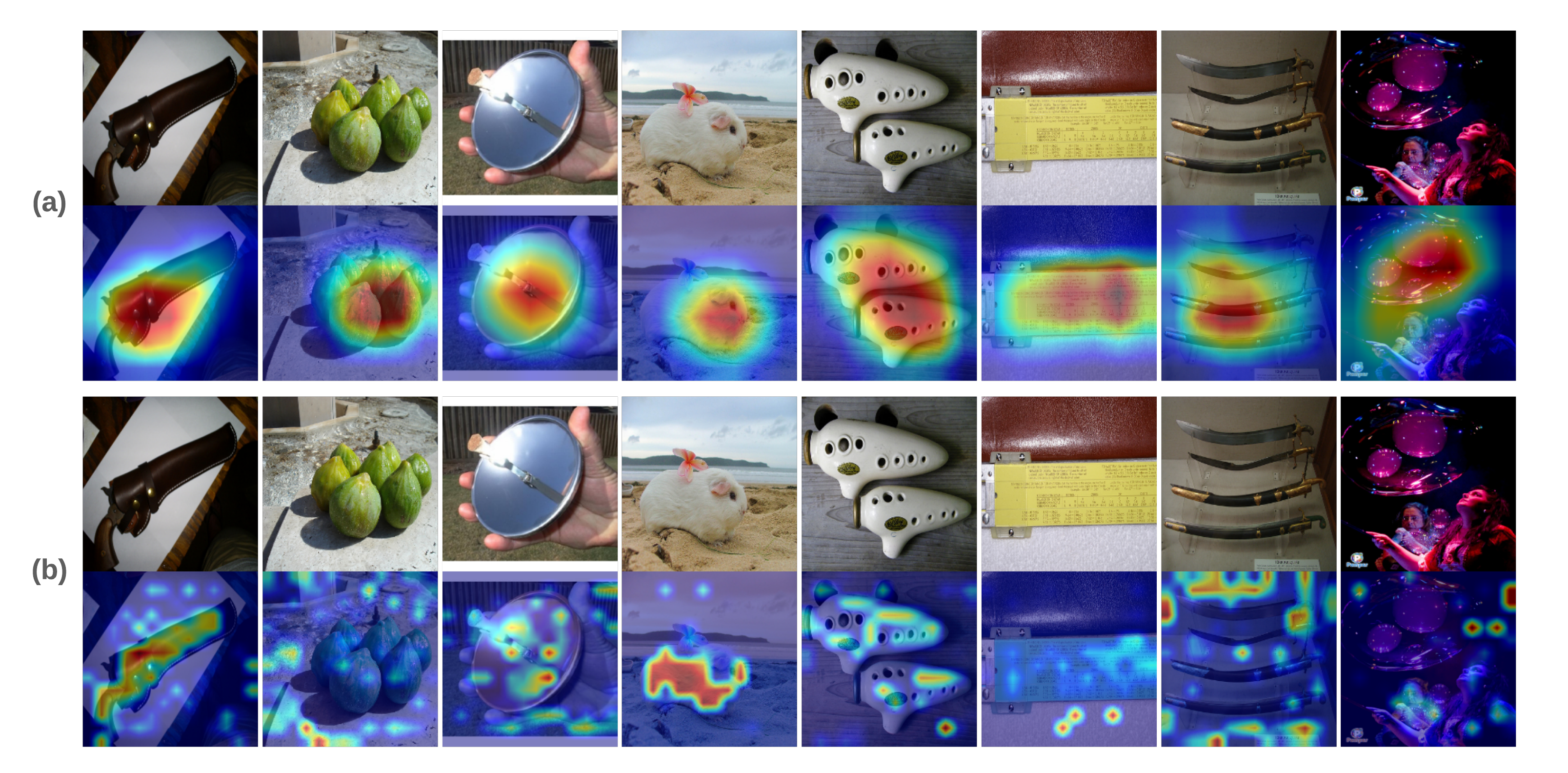} 
\caption{Additional Grad-CAM results. Predictions are (left-to-right) ``holster", ``fig", ``solar\_dish", ``guinea\_pig", ``ocarina", ``slide\_rule", ``scabbard", and ``bubble". All the predictions are correct. (\textbf{a}) BiT (\texttt{m-r101x3}). (\textbf{b}) ViT L-16.}
\label{fig:gradcam-more}
\end{figure}

\section{Additional Results on ImageNet-C}
\paragraph{Scores on individual corruptions of ImageNet-C.}

In Table \ref{tab:indv-top-1-imagenet-c}, we provide the individual top-1 accuracy scores for the 15 different corruption types of ImageNet-C. Note that for this we only consider the severity level of 5. As mentioned in Section \ref{para:imagenet-c}, ViT particularly performs poorly on the "contrast" corruption. 

\begin{table*}[h]
\centering
\resizebox{\textwidth}{!}{%
\begin{tabular}{@{}cccccccccccccccc@{}}
\toprule
 & \multicolumn{3}{c}{\textbf{Noise}} & \multicolumn{4}{c}{\textbf{Blur}} & \multicolumn{4}{c}{\textbf{Weather}} & \multicolumn{4}{c}{\textbf{Digital}} \\ \midrule
Model        & Gauss & Shot  & Impulse & Defocus & Glass & Motion & Zoom  & Snow  & Frost & Fog   & Bright & Contrast                     & Elastic & Pixelate & JPEG  \\
BiT-m r50x1  & 26.25 & 26.81 & 27.65   & 21.7    & 12.44 & 25.94  & 26.32 & 28.24 & 36.38 & 27.77 & 58.94  & 16.88                        & 19.08   & 49.22    & 49.77 \\
BiT-m r50x3  & 35.96 & 36.69 & 37.62   & 31.56   & 16.7  & 30.42  & 31.93 & 30.65 & 43.02 & 31.89 & 64.35  & 21.42                        & 22.39   & 53.83    & 57.43 \\
BiT-m r101x1 & 31.6  & 32.58 & 33.16   & 26.34   & 16.59 & 29     & 29.82 & 28.53 & 37.56 & 28.03 & 60.43  & 16.07                        & 21.39   & 48.79    & 52.91 \\
BiT-m r101x3 & 36.25 & 36.48 & 38.18   & 33.8    & 16.25 & 38.52  & 36.9  & 36.51 & 43.81 & 26.64 & 64.56  & 17.34                        & 27.48   & 56.88    & 56.08 \\
BiT-m r152x4 & 35.08 & 35.38 & 37.44   & 34.66   & 18.66 & 35.75  & 39.82 & 33.98 & 43.42 & 31.03 & 63.62  & 24.38                        & 27.13   & 56.66    & 55.35 \\
\multicolumn{16}{c}{}                                                                                                                                                  \\
ViT B-16     & 23.27 & 26.32 & 25.43   & 31.65   & 28.8  & 43.36  & 39.85 & 46.1  & 45.99 & 42.05 & 71.96  & {\color[HTML]{FE0000} 12.93} & 41.91   & 60.77    & 57.52 \\
ViT B-32     & 23.45 & 25.43 & 25.39   & 31.71   & 28.97 & 38.39  & 33.67 & 34.83 & 42.52 & 33.44 & 68     & {\color[HTML]{FE0000} 9.31}  & 44.71   & 60.8     & 54.79 \\
ViT L-16     & 39.1  & 39.95 & 42.1    & 36.71   & 36.3  & 49.69  & 47.65 & 52.82 & 51.53 & 47.99 & 74.53  & {\color[HTML]{FE0000} 18.85} & 49.87   & 70.05    & 63.45 \\
ViT L-32     & 33.56 & 35.06 & 35.07   & 36.06   & 33.86 & 44.5   & 40.73 & 39.53 & 45.54 & 38.28 & 68.93  & {\color[HTML]{FE0000} 8.76}  & 49.94   & 62.17    & 59.31 \\ \bottomrule
\end{tabular}%
}
\vspace{-2mm}
\caption{Individual top-1 accuracy scores (\%) on all the corruption types of ImageNet-C.}
\label{tab:indv-top-1-imagenet-c}
\end{table*}

Next, in Table \ref{tab:indv-ce}, we report the individual unnormalized corruption errors (not scaled using the AlexNet errors) on the same 15 different corruptions as given by BiT-m \texttt{r101x3} and ViT L-16. 

\begin{table*}[h]
\centering
\resizebox{\textwidth}{!}{%
\begin{tabular}{@{}cccccccccccccccc@{}}
\toprule
\multicolumn{1}{l}{} &
  \multicolumn{3}{c}{\textbf{Noise}} &
  \multicolumn{4}{c}{\textbf{Blur}} &
  \multicolumn{4}{c}{\textbf{Weather}} &
  \multicolumn{4}{c}{\textbf{Digital}} \\ \midrule
Model        & Gauss & Shot  & Impulse & Defocus & Glass & Motion & Zoom  & Snow  & Frost & Fog   & Bright & Contrast & Elastic & Pixelate & JPEG  \\
BiT-m \texttt{r101x3} & 44.45 & 45.88 & 46.7    & 48.18   & 64.34 & 44.7   & 50.03 & 51.18 & 46.3  & 51.48 & 28.48  & 45.49    & 49.3    & 34.31    & 36.53 \\
ViT L-16     & 35.05 & 35.98 & 35.88   & 41.78   & 42.45 & 34.59  & 40.73 & 40.47 & 40.42 & 39.61 & 22.64  & 43.57    & 32.97   & 24.31    & 27.79 \\ \bottomrule
\end{tabular}%
}
\vspace{-2mm}
\caption{Individual unnormalized corruption errors (\%) on 15 different corruption types of ImageNet-C.}
\label{tab:indv-ce}
\end{table*}

\paragraph{Stability under common corruptions.}

\begin{figure}[ht]
\centering
\includegraphics[width=1.0\columnwidth]{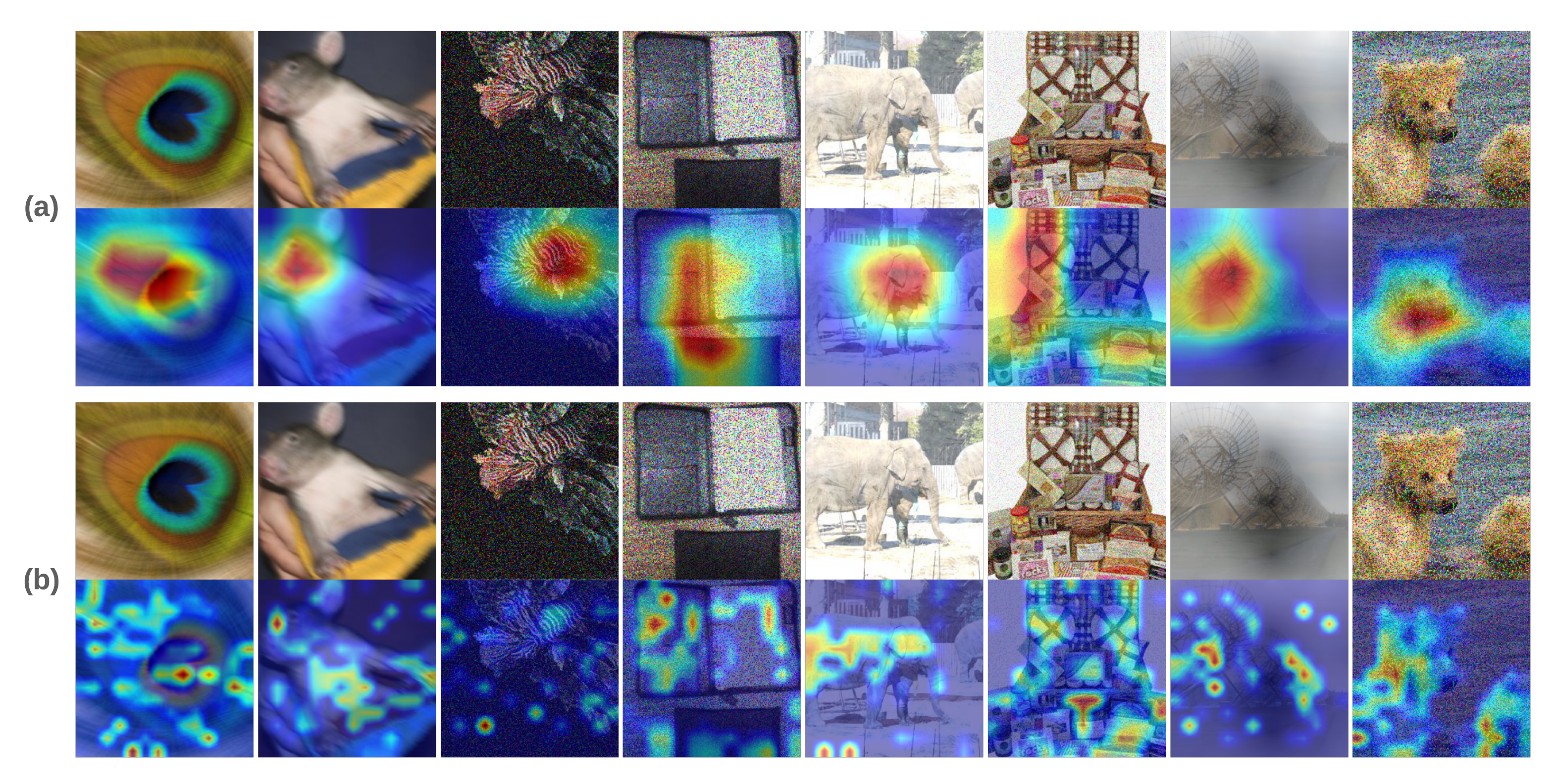} 
\caption{Grad-CAM results on ImageNet-C. For a fair comparison, we only consider the case where BiT makes wrong predictions whereas ViT predictions are still correct. The predictions (with confidence scores) are (left-to-right): ``quill" (98.02\%), ``wombat" (92.54\%), ``lionfish" (99.55\%), ``binder"  (80.29\%), ``Indian\_elephant" (59.60\%), ``hamper" (99.82\%), ``radio\_telescope" (97.93\%), and ``brown\_bear" (79.02\%). (\textbf{a}) BiT (\texttt{m-r101x3}). (\textbf{b}) ViT L-16.}
\label{fig:gradcam-imagenet-c}
\end{figure}

As reported in Section \ref{para:imagenet-c}, ViT significantly outperforms BiT when exposed to common corruptions. To better understand if ViT is able to hold its attention-span under those corruptions, in Figure \ref{fig:gradcam-imagenet-c}, we provide Grad-CAM results for a few images from ImageNet-C sampled from different levels of severity.

While the Grad-CAM results for BiT may seem more appealing but the slight corruptions are enough to make its predictions flip. However, that is not the case for ViT. By utilizing more global context it is able to perform with strong confidence even under these corruptions. 

\section{Additional Results on ImageNet-P}

In Table \ref{tab:indv-fr-imagenet-p} and Table \ref{tab:indv-td-imagenet-p}, we report the flip rates and top-5 distances of the individual perturbation types of ImageNet-P respectively with BiT and ViT. These scores are unnormalized meaning that they were not scaled using corresponding AlexNet scores. 

\begin{table*}[h]
\centering
\resizebox{\textwidth}{!}{%
\begin{tabular}{@{}ccccccccccc@{}}
\toprule
\multicolumn{1}{l}{} & \multicolumn{2}{c}{Noise} & \multicolumn{2}{c}{Blur} & \multicolumn{2}{c}{Weather} & \multicolumn{4}{c}{Digital} \\ \midrule
Model        & Gauss  & Shot   & Motion & Zoom  & Snow  & Bright & Translate & Rotate & Tilt  & Scale \\
BiT-m r101x3 & 11.932 & 13.686 & 4.594  & 3.665 & 5.523 & 3.23   & 4.315     & 6.121  & 3.893 & 9.273 \\
ViT L-16     & 7.363  & 8.263  & 2.388  & 1.945 & 2.969 & 2.031  & 3.63      & 5.085  & 2.788 & 8.434 \\ \bottomrule
\end{tabular}%
}
\vspace{-2mm}
\caption{Individual unnormalized flip rates (\%) on all the perturbation types of ImageNet-P.}
\label{tab:indv-fr-imagenet-p}
\end{table*}

\begin{table*}[h]
\centering
\resizebox{\textwidth}{!}{%
\begin{tabular}{@{}ccccccccccc@{}}
\toprule
\multicolumn{1}{l}{} &
  \multicolumn{2}{c}{\textbf{Noise}} &
  \multicolumn{2}{c}{\textbf{Blur}} &
  \multicolumn{2}{c}{\textbf{Weather}} &
  \multicolumn{4}{c}{\textbf{Digital}} \\ \midrule
Model        & Gauss   & Shot    & Motion  & Zoom    & Snow    & Bright  & Translate & Rotate  & Tilt    & Scale   \\
BiT-m r101x3 & 3.8449  & 4.2462  & 1.44212 & 1.20941 & 1.75846 & 1.1756  & 1.59866   & 2.1012  & 1.45261 & 2.75042 \\
ViT L-16     & 2.49186 & 2.68482 & 0.7974  & 0.63717 & 0.92836 & 0.69184 & 1.27774   & 1.67676 & 1.0207  & 2.39877 \\ \bottomrule
\end{tabular}%
}
\vspace{-2mm}
\caption{Individual unnormalized top-5 distances (\%) on all the perturbation types of ImageNet-P.}
\label{tab:indv-td-imagenet-p}
\end{table*}

\section{Random Masking with Cutout}

\begin{figure}[ht]
\centering
\includegraphics[width=0.95\columnwidth]{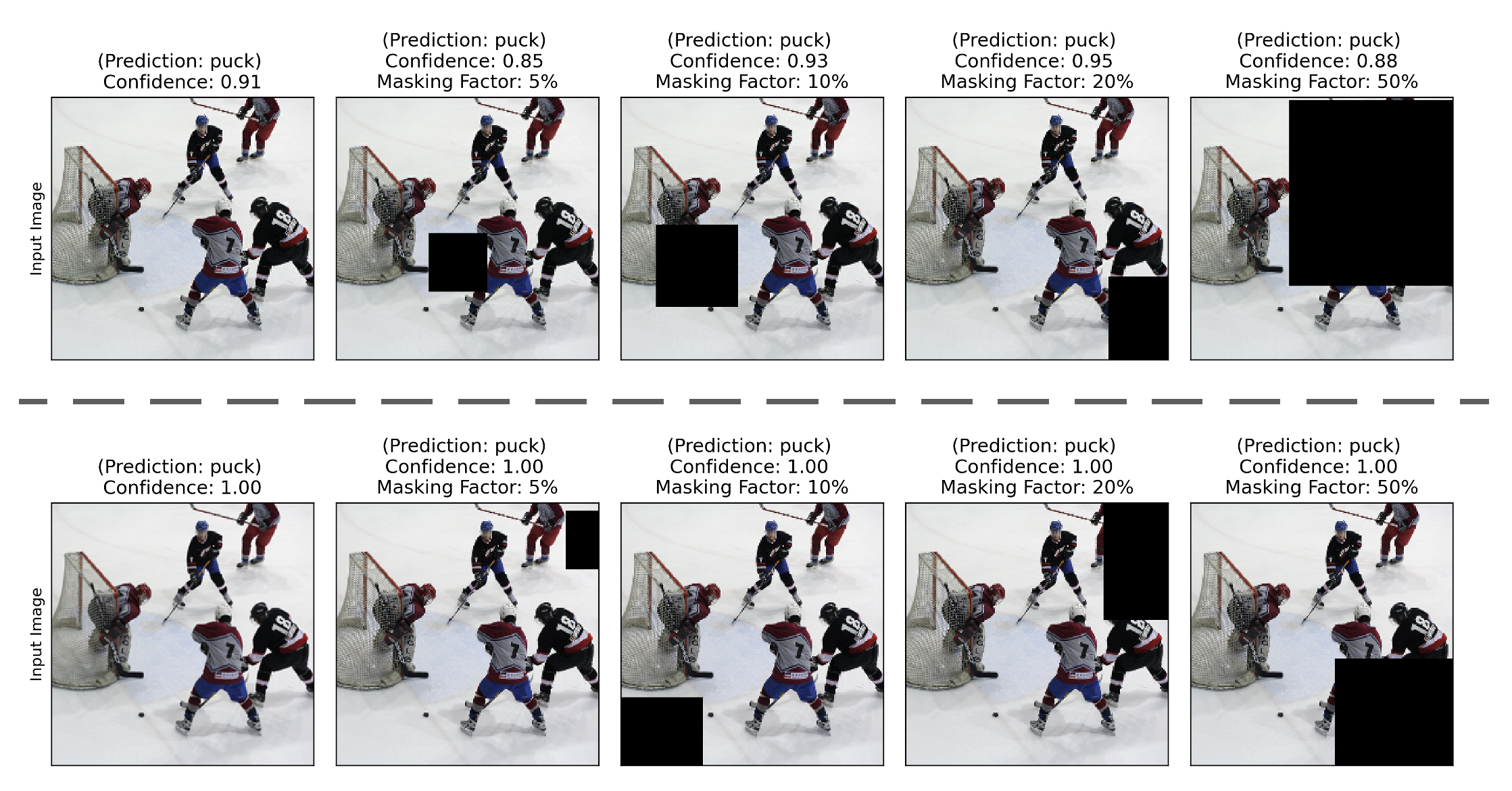} 
\caption{Change of prediction confidence as a function of masking. First row is from BiT-m \texttt{r101x3} and second row is from ViT L-16.}
\label{fig:cutout}
\end{figure}

In Figure \ref{fig:cutout}, we show how the predictions of BiT (\texttt{m-r101x3}) and ViT (L-16) change as a function of the masking factor in Cutout \cite{devries2017improved}. We provide these results for giving a clearer sense of the experiments conducted in Section \ref{masking-cutout}. These results should not be treated as anything conclusive. 

\section{Magnitude Spectrum and High-Frequency Components}

\begin{figure}[ht]
\centering
\includegraphics[width=0.95\columnwidth]{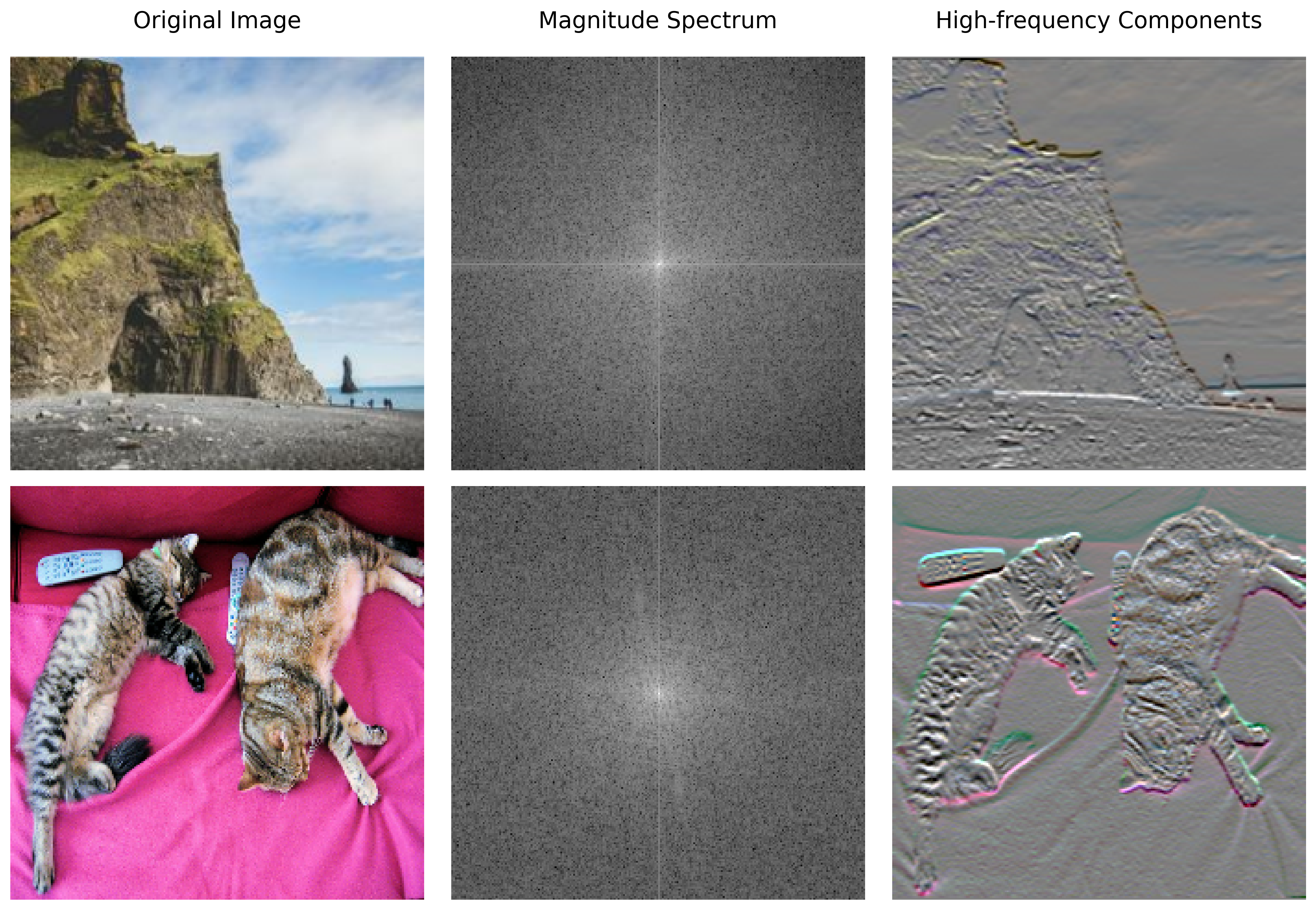} 
\caption{Visualization of the magnitude spectrum in the frequency domain and raw high-frequency components.}
\label{fig:magnitude-spec}
\end{figure}

Since we use Fourier analysis in Section \ref{fourier-analysis}, in the interest of comprehensiveness, we provide visualizations of the magnitude spectrum of frequency components as well as the raw high-frequency components of natural images in Figure \ref{fig:magnitude-spec}. 

\section{Adversarial Perturbations}
\paragraph{Peaking into the adversarial perturbations.}
\label{perturbations}

\begin{figure*}[h]
\centering
\includegraphics[width=0.95\textwidth]{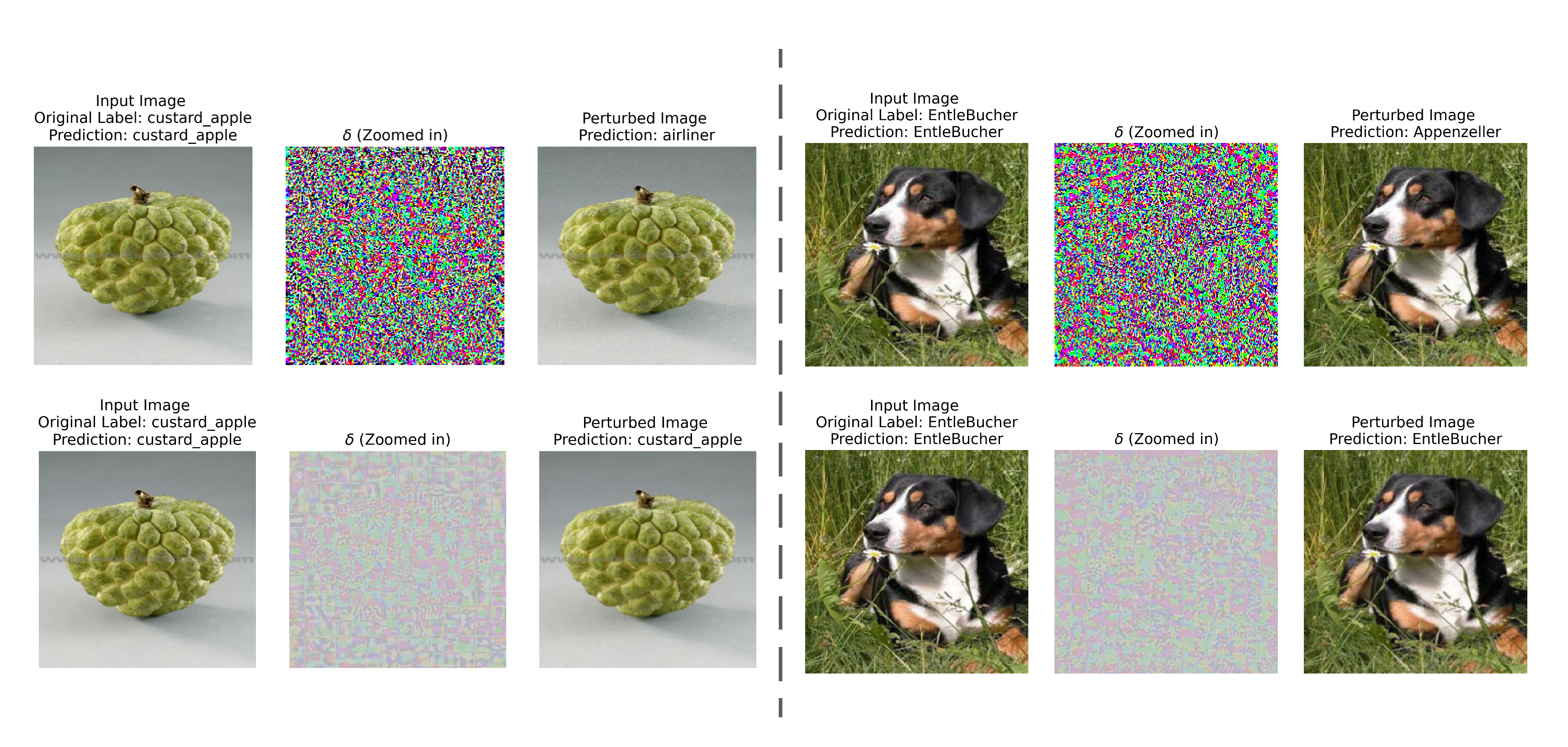} 
\caption{Visualization of the adversarial perturbations. First row is from BiT-m \texttt{r101x3} and second row is from ViT L-16.}
\label{fig:perturbations}
\end{figure*}

In Figure \ref{fig:perturbations}, we visualize the perturbations as learned by BiT-m \texttt{r101x3} and ViT L-16. We use Adam as optimizer here to implement PGD attack (see Section \ref{loss-perturbations} for implementation details) with a learning rate of 1e-3.   Generally, we find that the perturbations are smoother in case of ViT. 

\paragraph{Loss landscape of individual examples.}

\begin{figure*}[h]
\centering
\includegraphics[width=0.9\textwidth]{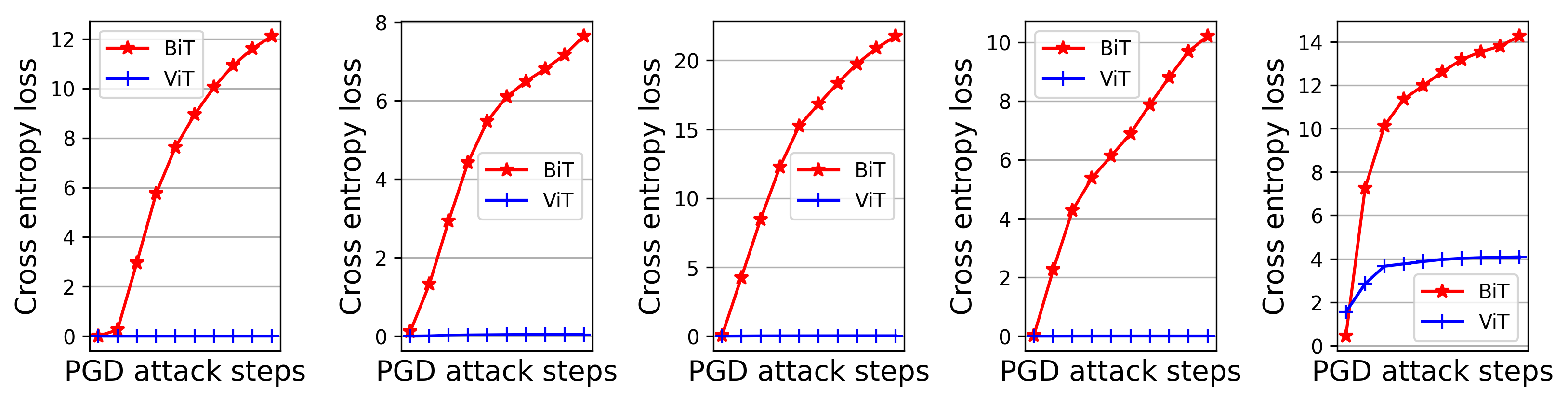} 
\caption{PGD loss plots of individual examples from the ImageNet-1k validation set.}
\label{fig:indv-loss}
\end{figure*}

In Figure \ref{fig:indv-loss}, we show PGD loss plots from five individual ImageNet-1k validation images used in Section \ref{loss-perturbations}. These examples are not cherry-picked and have been provided to better isolate the results shown in Figure \ref{fig:pgd}.

\end{document}